\documentclass{article}

\usepackage{arxiv}

\usepackage[utf8]{inputenc} 
\usepackage[T1]{fontenc}    
\usepackage{hyperref}       
\usepackage{url}            
\usepackage{booktabs}       
\usepackage{amsfonts}       
\usepackage{nicefrac}       
\usepackage{microtype}      
\usepackage{lipsum}
\usepackage{graphicx}

\usepackage{multicol}
\usepackage{comment}
\usepackage{cite}
\usepackage{amsmath,amssymb,amsfonts}
\usepackage{algorithmic}
\usepackage{graphicx}
\usepackage{textcomp}
\usepackage{multirow}
\usepackage{amsmath,amsfonts,amssymb}

\usepackage{verbatim}
\usepackage{mathtools}
\usepackage{upgreek}
\usepackage{booktabs}
\usepackage[dvipsnames]{xcolor}
\usepackage{footnote}
\makesavenoteenv{tabular}
\makesavenoteenv{table}

\usepackage{amsmath}
\usepackage{url}
\usepackage{subcaption} 

\usepackage{epsfig}
\usepackage{subcaption}

\usepackage{calc}
\usepackage{amssymb}
\usepackage{amstext}
\usepackage{amsmath}
\usepackage{amsthm}
\usepackage{multicol}
\usepackage{pslatex}
\usepackage{float}
\usepackage{gensymb}

\usepackage{multirow}
\usepackage{url}

\usepackage{breakurl}
\usepackage{hyperref}

\title{Impact of Thermal Throttling on Long-Term Visual Inference in a CPU-based Edge Device}

\author{
  Th\'{e}o Benoit-Cattin
  \\ \'{E}cole Nationale Sup\'{e}rieure d'\'{E}lectrotechnique, d'\'{E}lectronique,\\ d'Informatique, d'Hydraulique et des T\'{e}l\'{e}communications \\ Toulouse, France
  \And
  Delia~Velasco-Montero 
  \\ Instituto de Microelectr\'{o}nica de Sevilla \\ Universidad de Sevilla-CSIC
  \And
  Jorge~Fern\'{a}ndez-Berni 
    \\ Instituto de Microelectr\'{o}nica de Sevilla \\ Universidad de Sevilla-CSIC
}

\date{}
\renewcommand{\headeright}{}
\renewcommand{\undertitle}{}

\begin{document}
\maketitle
\begin{abstract}
Many application scenarios of edge visual inference, e.g., robotics or environmental monitoring, eventually require long periods of continuous operation. In such periods, the processor temperature plays a critical role to keep a prescribed frame rate. Particularly, the heavy computational load of convolutional neural networks (CNNs) may lead to thermal throttling and hence performance degradation in few seconds. In this paper, we report and analyze the long-term performance of 80 different cases resulting from running 5 CNN models on 4 software frameworks and 2 operating systems without and with active cooling. This comprehensive study was conducted on a low-cost edge platform, namely Raspberry Pi 4B (RPi4B), under stable indoor conditions. The results show that hysteresis-based active cooling prevented thermal throttling in all cases, thereby improving the throughput up to approximately 90\% versus no cooling. Interestingly, the range of fan usage during active cooling varied from 33\% to 65\%. Given the impact of the fan on the power consumption of the system as a whole, these results stress the importance of a suitable selection of CNN model and software components. To assess the performance in outdoor applications, we integrated an external temperature sensor with the RPi4B and conducted a set of experiments with no active cooling in a wide interval of ambient temperature, ranging from 22 ºC to 36 ºC. Variations up to 27.7\% were measured with respect to the maximum throughput achieved in that interval. This demonstrates that ambient temperature is a critical parameter in case active cooling cannot be applied. 
\end{abstract}

\keywords{ Edge Vision \and Long-Term Inference \and Thermal Throttling \and Convolutional Neural Networks \and Raspberry Pi \and Ambient Conditions }

\section{Introduction}
\label{sec:introduction}

 Deep learning (DL) \cite{LeCun2015} and its particular embodiment in the form of convolutional neural networks (CNNs) have become the de-facto approach to address many computer vision tasks, e.g., image recognition, object detection, or segmentation. New training and processing techniques together with the availability of large datasets and high computational power are the main underlying factors supporting the distinctive feature of CNNs versus classical algorithms, i.e., their high accuracy. The flip side of this high accuracy is a notable increase in processing and memory requirements, which has prompted multiple research efforts on designing efficient network architectures \cite{Googlenet_7298594,shufflenet_inproceedings,shufflenetv2,Mobilenet_DBLP:journals/corr/HowardZCKWWAA17,MobilenetV2_8578572,Squeezenet_DBLP:journals/corr/IandolaMAHDK16,Efficientnet_pmlr-v97-tan19a} 
and reducing the computational load of CNNs through techniques such as network pruning, data quantization, and network compression \cite{IncrementalNQ,PrunningCNNfor_DBLP:conf/iclr/MolchanovTKAK17,Kim2015Compression_DBLP:journals/corr/KimPYCYS15}.

Despite these efforts, the implementation of CNNs still constitutes a major challenge for edge visual inference. Application scenarios such as robotics \cite{fumebot}, environmental monitoring \cite{Fern12}, or the Internet of Things (IoT) \cite{IoT_lin17} can greatly benefit from incorporating vision capabilities in ultra-low-power low-cost devices. However, DL-based processing pipelines deplete the scarce resources available in such devices, significantly affecting the timely completion of other tasks. Furthermore, CNNs can degrade their own performance -- and therefore the performance of the whole system -- as a result of thermal throttling, precluding continuous inference at prescribed throughput during a long runtime period. 

In this study, we expressly focused on the problem of thermal throttling, i.e., the automatic reduction of processor performance caused by temperature excess. It is a preventive action to avoid thermal damage of system components and implies a reduction of the clock frequency. Our analysis is based on an extensive set of indoor measurements comprising state-of-the-art CNN models and software frameworks deployed on the latest version of arguably the most popular embedded CPU platform, i.e., Raspberry Pi 4B (RPi4B). In addition to the standard 32-bit operating system (OS) usually running on RPi4B, we also characterized the 64-bit beta version recently released. This vast set of measurements is intrinsically valuable as a basis to compare and select the optimal combination of components according to prescribed specifications for edge vision. Concerning thermal throttling, we examined the long-term performance of 80 combinations without and with active cooling in terms of throughput; CPU usage, temperature, and frequency; and fan usage. The best combinations for each parameter are pointed out and interesting conclusions are drawn. For instance, we prove that the 64-bit OS is not always the best option, in spite of the fact that it is supposed to better exploit the 64-bit architecture of the RPi4B CPU. We also analyze the case of outdoor operation to elucidate the impact of varying ambient temperature with no active cooling. This is critical for remote sensing applications in which the extra power consumption of the fan could not be affordable. 
 
The remainder of this paper is organized as follows. In Section~\ref{sec:realtedwork}, related studies reported in the literature are briefly reviewed and our contribution is emphasized. The experimental setup for assessing thermal throttling during CNN inference together with the evaluated hardware and software components are introduced in Section~\ref{sec:preliminaries}. The characterization methodology is described in Section~\ref{sec:benchmark}. The indoor experimental results are presented and discussed in Section~\ref{sec:results} whereas the outdoor case is addressed in Section~\ref{sec:temp_experiment}. Finally, we draw relevant conclusions in Section~\ref{sec:conclusion}.

\section{Related Work}
\label{sec:realtedwork}

 Throughput, energy budget, or memory restrictions are major concerns when designing an embedded system for edge vision. These days, it implies to carefully consider the heterogeneity of CNN models and specialized hardware and software available for DL-based inference. Regarding the ever-growing zoo of CNN models, comparative studies facilitate model selection according to the prescribed application accuracy and model size suitable for the target platform \cite{An_Analysis_of_DNN_CanzianiPC16,Benchmark_analysis}. However, there is a strong dependence on the underlying hardware integration and software framework. In recent studies, the inference time and energy consumption of CNNs on various edge platforms and hardware accelerators were measured \cite{ostrowski,mlmark,AI_benchmark}. At software level, the performance of a diversity of frameworks for mobile applications was also evaluated in \cite{luo2020comparison}, where it is stated that the software stack contributes to the inference performance in the same proportion as model complexity does. The recent project \textit{MLPerf} \cite{mlperf_10.1109/ISCA45697.2020.00045} constitutes an attempt to address this complex scenario holistically. It is a collaborative benchmark framework to provide reliable performance figures for multiple models, hardware platforms, and software environments. 

In view of these benchmarking efforts, it is remarkable that thermal issues derived from running CNNs on edge devices have hardly been addressed in previous works.  In \cite{Choi_AnAdaptiveandIntegrate}, different algorithms were studied for managing both platform overheating and power consumption during extensive workloads on IoT devices. To this end, the authors applied well-known techniques such as dynamic power management and dynamic voltage and frequency scaling (DVFS). However, they focused on basic vision tasks and did not analyze the impact of thermal throttling. Two power control policies for embedded systems were evaluated in \cite{peluso_et_al_2}. CNNs were characterized under DVFS thermal management based on such policies. In a more recent study \cite{peluso_et_al_1}, the same authors analyzed the impact of model topology scaling on power-constrained systems over both functional (accuracy) and non-functional (latency and temperature) constraints. In both works \cite{peluso_et_al_2,peluso_et_al_1}, the authors stressed the importance of thermally induced performance degradation and the scarcity of studies on the topic. 

In this context, the first major contribution of our study is a thorough assessment of the impact of thermal throttling on a broad set of state-of-the-art CNN workloads running on a low-cost CPU platform under stable ambient temperature. Performance metrics are compared with the case of hysteresis-based active cooling, which prevented performance degradation for all the workloads. A second contribution is the evaluation of fan usage during active cooling, which allows to quantify the extra power consumption required to keep maximum performance; interestingly, the fan usage widely varied over the benchmarking set. Finally, a third important contribution is the analysis of outdoor performance in a wide range of ambient temperature with no cooling. We consider that the discussion and conclusions about our results will also be of interest for the related research community. 

\section{Experimental Setup}
\label{sec:preliminaries}

\begin{figure}[!h]
\centering
\includegraphics[width=0.4\textwidth]{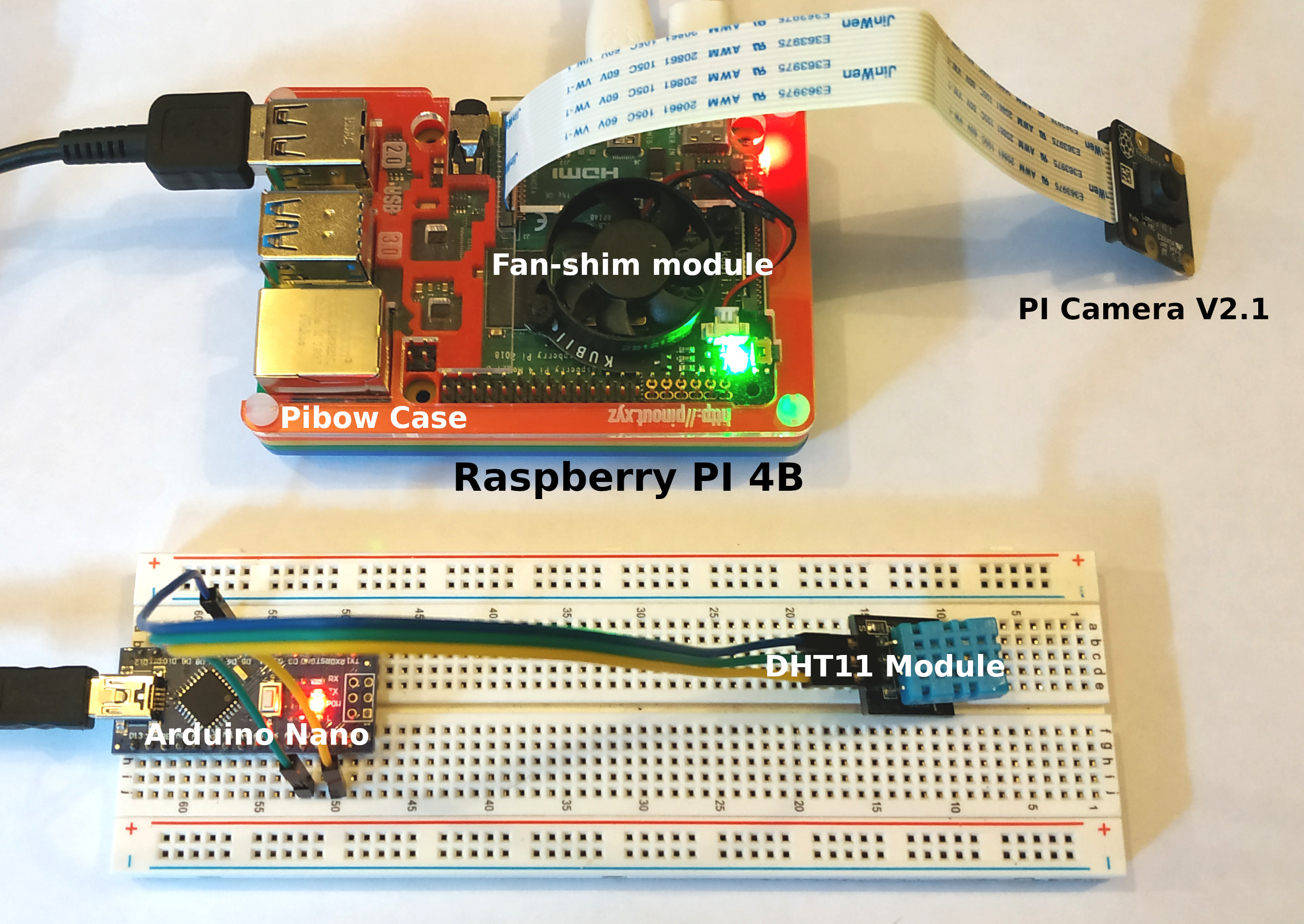}
\caption{Experimental setup: visual inference was performed on the RPi4B, which also processed temperature values from a DHT11 sensor controlled by an Arduino Nano microcontroller.} \label{fig:setup}
\end{figure}

Figure \ref{fig:setup} shows the experimental setup employed for the different characterizations performed in this study. The central device is the RPi4B. An off-the-shelf camera module \cite{RPi-Camera} was also integrated, but it was not finally used for testing. Instead, images previously stored in memory provided a constant input flow, thereby ensuring that no bias on performance occurred due to changes in illumination conditions. The ambient temperature was sensed by a DHT11 temperature sensor \cite{DHT11-Datasheet} connected to an Arduino Nano microcontroller, which transmits the temperature values to the RPi4B through the serial port. Next, we provide further details about the RPi4B platform and briefly describe the different components of the software stack and the evaluated CNN models.

\subsection{Hardware Platform}

CNN-based computer vision was implemented on the RPi4B \cite{raspberryPi4}. Its Broadcom BCM2711 system-on-chip includes a quad-core ARMv8  Cortex-A72 64-bit CPU that can work at frequencies ranging from 600 MHz to 1.5 GHz. It also incorporates a Broadcom VideoCore VI GPU, but it cannot be easily exploited for CNN inference at the present time -- i.e., leveraging open-source DL software for GPU acceleration. We specifically used a RPi4B model featuring 8 GB LPDDR4-3200 SDRAM.

Heavy processing workloads resulting from running CNNs cause CPU overheating, which is automatically counteracted through a DVFS strategy encoded in the RPi4B firmware \cite{rpi_freq_management}. When the temperature of the quad-core ARM processor exceeds a critical value ($T_{limit}=80$ \degree C for RPi4B), the clock frequency of the cores is progressively reduced from 1.5 GHz to, eventually, 600 MHz, thereby gradually degrading CNN inference performance. Alternatively, CPU overheating can be mitigated through an external cooling fan. We employed a commercial module \cite{Fan-shim-page} compatible with RPi4B consisting of a fan unit mounted on a circuit board. This board allows the user to control the fan with a hysteresis approach in accordance with the CPU temperature through a Python library \cite{Fan-shim-python}. The fan consumes 0.65 W when it is on. Given that the power consumption of the RPi4B ranges from 3 W (idle) to 6.25 W (under stress) \cite{rpi_faqs}, the fan increases the power consumption of the system at least by approximately 10\%. 

\subsection{Software Stack}
\subsubsection{Operating System}
 Raspberry Pi Foundation officially provides a 32-bit OS for RPi4B -- so-called Raspbian \cite{raspberryPiOS-32bit}.
Nonetheless, a 64-bit beta version of this OS was recently released \cite{raspberryPiOS-64bit}. This second version aims to completely leverage the 64-bit ARM processor of the platform. Experiments were conducted on both Raspbian flavors. 

\subsubsection{DL Frameworks}

 Training, testing, and deployment of DL-based applications are facilitated by DL software frameworks. Within the diversity of tools publicly available these days, we selected the following frameworks specifically oriented for embedded inference: \\

\noindent \textit{OpenCV} \cite{OpenCV} is a popular library that allows importing DNN models previously trained with other tools -- e.g., Caffe, TensorFlow, or Torch -- to perform visual inference. We built OpenCV version 4.3.0-dev by setting the pertinent compilation flags to exploit the ARM NEON instruction set and enable VFPv3 optimizations. This library was accessed through its C++ interface.  \\

\noindent \textit{Tengine} \cite{Tengine}, developed by OPEN AI LAB \cite{OPENAILAB}, is a tool intended for running neural networks on embedded devices and IoT scenarios. It is specifically designed to make the most of ARM architectures. Pre-trained models from Tensorflow, Caffe, and MXNet, in addition to ONNX models, can be converted to the particular model format employed by Tengine. We used the standalone version of Tengine v1.12.0 and C++ coding. \\

\noindent \textit{NCNN}  \cite{NCNN} is an inference framework optimized for mobile platforms. Its high-performance implementation, exclusively encoded in C++, supports ARM NEON optimizations and multi-core parallel computing acceleration. It can import models from Caffe, PyTorch, MXNet, ONNX, and DarkNet.\\

\noindent \textit{ArmNN SDK} \cite{ARMNN} is a set of open-source Linux software tools that enables machine leaning workloads on power-efficient devices. It provides a bridge between existing neural network frameworks and low-power Cortex-A CPUs, ARM Mali GPUs, and ARM Ethos neural processing units. Consequently, it is supposed to fully leverage the RPi4B cores. This software can take networks from other DL frameworks, translate them into the internal ArmNN format and deploy them efficiently. It supports Caffe, TensorFlow, TensorFlow Lite, and ONNX models. The version of ArmNN we used is v20.05 through its Python API (pyArmNN).\\

\subsection{Convolutional Neural Networks}
\label{CNNs}

We benchmarked the five CNNs listed in Table \ref{tab:CNNs}. All of them were trained over the ImageNet dataset \cite{ImageNet_CVPR} for 1000-category image classification. The set includes complex -- ResNets -- and lightweight -- MobileNets and SqueezeNet -- models. Note that there is no clear correlation between an increasing number of weights and a corresponding higher accuracy. This has a significant impact on the achievable performance, as will be demonstrated in Section~\ref{sec:results}.  We downloaded the files with the corresponding pre-trained network weights compatible with each framework from the public repositories shown in Table \ref{tab:RefModels}. 

\begin{table}[h!]
\caption{Popular CNNs for 1000-category image classification assessed in this study.}
\label{tab:CNNs} 
\centering
\scalebox{0.99}{
\begin{tabular}{l c l l l}
  \toprule
  & \textbf{Work} & \textbf{Top-1} & \textbf{Top-5} & \textbf{\#weights} \\ 
  \midrule
  \textbf{SqueezeNet-v1.1} & \cite{Squeezenet_DBLP:journals/corr/IandolaMAHDK16}& 57.5\% & 80.3\% & ~1.2 M \\
  \textbf{MobileNet-v1} & \cite{Mobilenet_DBLP:journals/corr/HowardZCKWWAA17} & 70.6\% & 89.9\% & ~4.2 M \\ 
  \textbf{MobileNet-v2} & \cite{MobilenetV2_8578572} & 72.0\% & 90.5\% & ~3.4 M \\ 
    \textbf{ResNet-18} & \cite{ResNet_7780459} & 69.1\% & 89.0\% & ~11.7 M \\
  \textbf{ResNet-50} & \cite{ResNet_7780459} & 77.2\% & 93.3\% & ~25.6 M \\
  \bottomrule
\end{tabular}}
\end{table}


\begin{table}[h!]
\caption{Repositories serving the files with pre-trained network weights used in this study. These files were either directly loaded for inference or translated to the format required by each framework.}
\label{tab:RefModels} 
\centering
\scalebox{0.99}{
\begin{tabular}{l c c c}
  \toprule
  &  \textbf{OpenCV} /  \textbf{Tengine} & \textbf{NCNN} & \textbf{ArmNN} \\
   \midrule
   \textbf{SqueezeNet-v1.1} & \cite{SqueezeNet-caffe} & \cite{NCNN-models} &  \cite{SqueezeNet-TFlite} \\
    \textbf{MobileNet-v1} &  \cite{Mobilenetv1-caffe} & \cite{NCNN-models} & \cite{MobileNetv1-TFlite} \\
    \textbf{MobileNet-v2} & \cite{Mobilenetv2-caffe} & \cite{NCNN-models}  & \cite{MobileNetv2-TFlite} \\
    \textbf{ResNet-18} & \cite{resnet18-caffe} & \cite{NCNN-models}  & \cite{ResNet-18-Keras} \\ 
   \textbf{ResNet-50} & \cite{resnet50-caffe} & \cite{NCNN-models}  &  \cite{ResNet-50-Keras} \\
  \bottomrule
\end{tabular}}
\end{table}

\section{Characterization Methodology}
\label{sec:benchmark}

 The long-term performance of each and every combination among the aforementioned OSs, software frameworks, and CNN models was evaluated on the RPi4B according to the methodology described next. 

\subsection{Metrics}
\label{sec:methodology}

Key performance metrics such as throughput, CPU usage, CPU temperature, and CPU frequency were periodically monitored during continuous inference periods. The experiments were repeated for two cases: 

\begin{itemize}
\item \textbf{case (a)}: raw version of the RPi4B hardware, i.e., no active cooling.
\item \textbf{case (b)}: application of active cooling through the external fan unit.
\end{itemize} 

A C++ test program was coded to extract the metrics\footnote{In the case of ArmNN, the test program was coded in Python given that we used the Python interface of this framework.}. This program proceeded as follows:

\begin{enumerate}
\item Images from ImageNet were resized to fit the input resolution of the pre-trained CNNs, i.e., $3\times224\times224$. This pre-processing time was not included in the throughput.
\item Real-time CNN inference with batch size $1$ was performed and the instantaneous throughput was measured.
\item  CPU usage, temperature, and frequency were monitored through calls to Linux tools, in particular \texttt{vcgencmd} and \texttt{mpstat}.
\end{enumerate}

The instantaneous ambient temperature was also periodically recorded during the tests. For the indoor experiments, this information allowed us to ensure that all the tests were conducted at ambient temperatures as stable and similar as possible, always within the range 27 -- 29 \degree C. For the outdoor experiments, the ambient temperature was correlated with the system performance under no-cooling conditions to evaluate the impact of the former on the latter. 

Finally, we also evaluated the fan usage, denoted by $FU$, for case (b) according to the following equation:

\begin{equation}
    FU (\%)=\frac{t_{ON}}{t_{T}} \times 100
\label{eq:fan}
\end{equation}
where $t_{ON}$ denotes the sum of time intervals in which the fan was on within the total test time, denoted by $t_{T}$. Note that when applying active cooling, a hysteresis approach is typically employed to reduce $FU$ as much as possible. We set the hysteresis on/off temperatures at 75 \degree C and 65 \degree C, respectively, to prevent thermal throttling on the RPi4B CPU.

\subsection{Tests}

\begin{figure}[!h]
\centering
\includegraphics[width=0.69\textwidth]{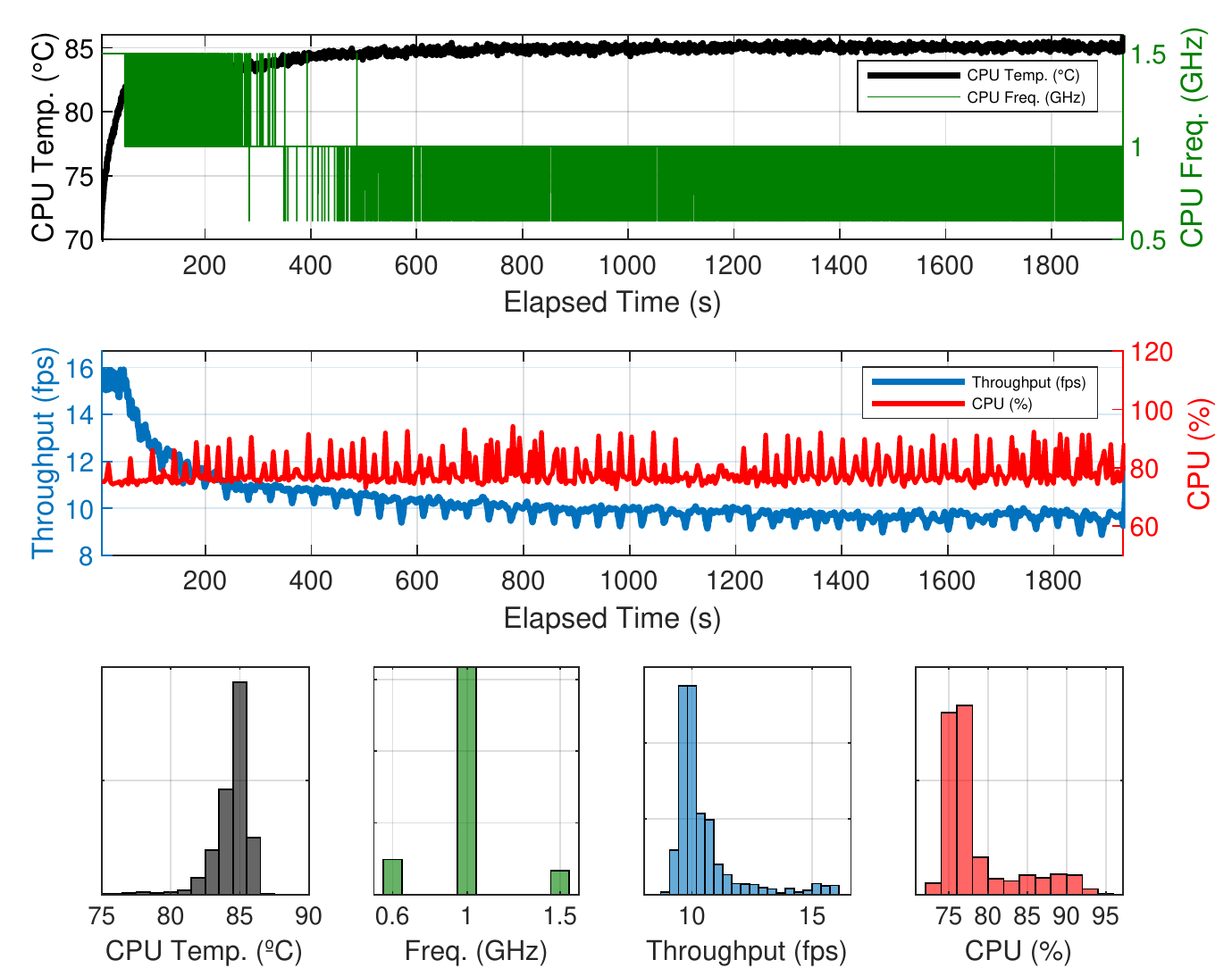}
\caption{Example of test for case (a) -- no cooling -- and 64-bit-OS/OpenCV/SqueezeNet combination. \textit{Top plot}: increasing CPU temperature leads to thermal throttling, forcing the reduction of the CPU frequency. \textit{Center plot}: throughput decreases due to thermal throttling, eventually becoming stable; CPU usage is also represented. \textit{Bottom plot}: histograms summarizing the plots above.} \label{fig:Example_Bare}
\end{figure}

\begin{figure}[!h]
\centering
\includegraphics[width=0.69\textwidth]{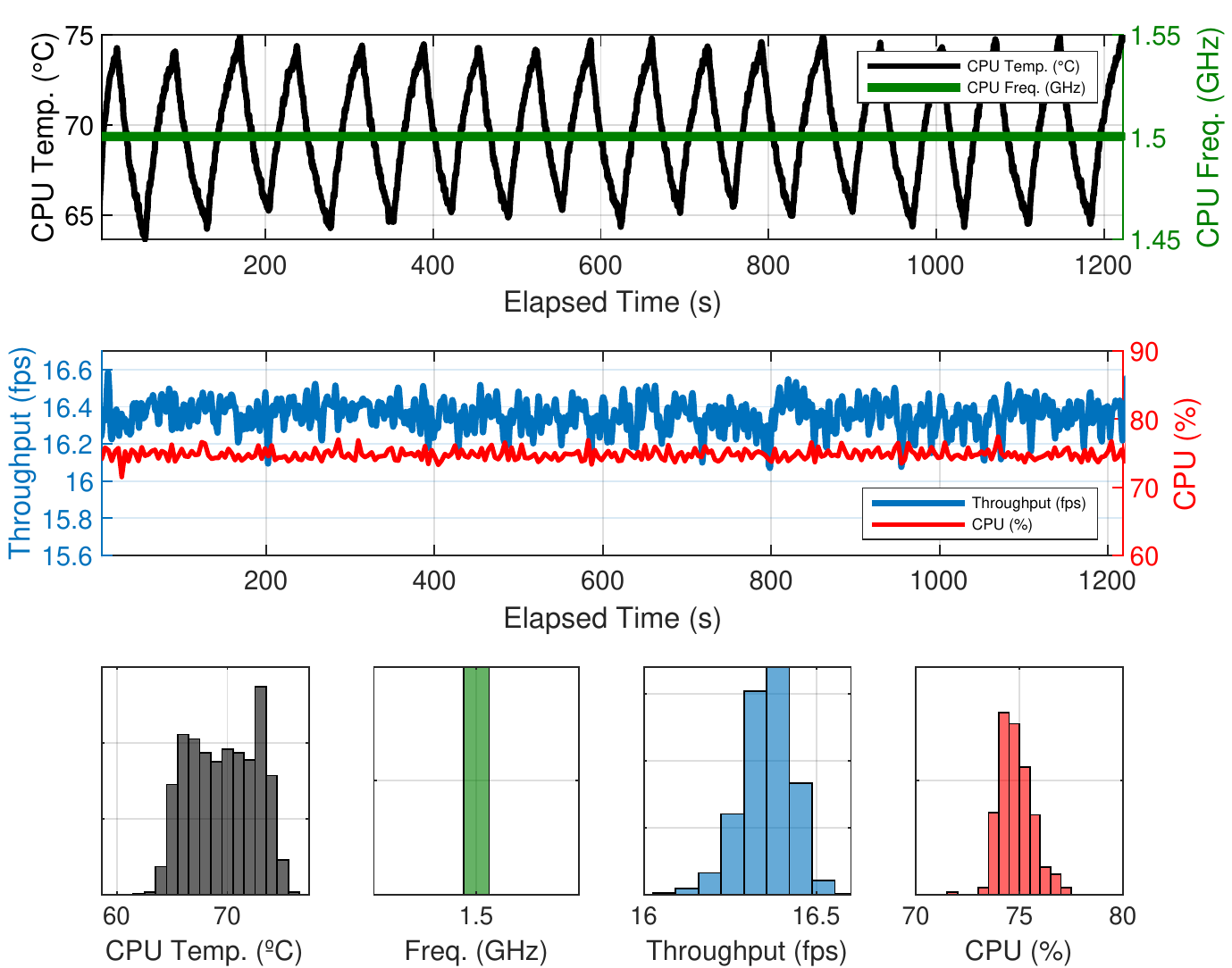}
\caption{Example of test for case (b) -- active cooling -- and 64-bit-OS/OpenCV/SqueezeNet combination. \textit{Top plot}: the CPU temperature clearly benefits from the fan hysteresis cycle, keeping the CPU frequency stable, in contrast with the top plot in Fig.~\ref{fig:Example_Bare}. \textit{Center plot}: stable throughput and CPU usage are achieved, in contrast with the center plot in Fig.~\ref{fig:Example_Bare}. \textit{Bottom plot}: histograms summarizing the plots above.} \label{fig:Example_Cooling}
\end{figure}

Each individual experiment was started when the CPU temperature was cool and stable, far away from thermal throttling. The two considered cases -- (a) no cooling; (b) active cooling -- were analyzed by averaging the aforementioned metrics at steady state. In case (a), steady state means that CPU overheating gave rise to thermal throttling and the DVFS strategy encoded in the RPi4B achieved a stable system performance. In case (b), active cooling maintained a steady performance throughout the total duration of all the tests. We exemplify these behaviours in Figs. \ref{fig:Example_Bare} and \ref{fig:Example_Cooling} for 64-bit OS/OpenCV/SqueezeNet.

Figure~\ref{fig:Example_Bare} shows the long-term behaviour of CNN inference without active cooling. The CPU temperature reached $T_{limit}$ after approximately 50 seconds of workload, causing thermal throttling and hence reduction of the CPU frequency from that instant on. Consequently, the throughput was notably affected, eventually reaching a stable value of approximately 10 fps, which is significantly lower than its initial value, around 16 fps. This occurred after approximately 1000 seconds of continuous inference, when the underlying DVFS strategy achieved a stable temperature around 85 \degree C. The histograms in the bottom plot reveal that the processor operated at 1 GHz most of the time, with peaks of CPU usage reaching 95\%.  


Figure \ref{fig:Example_Cooling} depicts the same CNN inference but applying active cooling. The impact of the fan hysteresis cycle on the CPU temperature is evident. Thermal throttling never occurred. Both CPU frequency and usage kept stable at 1.5 GHz and approximately 75\%, respectively. Likewise, the throughput slightly fluctuated around 16.4 fps. The histograms in the bottom plot confirm this steady behavior. Note that the CPU temperature swept a wider interval than in Fig.~\ref{fig:Example_Bare} but always below $T_{limit}$. 

%
%

\section{Indoor Tests: Results and Analysis}
\label{sec:results}

\begin{figure*}[!b]
     \centering
     \begin{subfigure}[b]{0.49\textwidth}
         \centering
         \includegraphics[width=\textwidth]{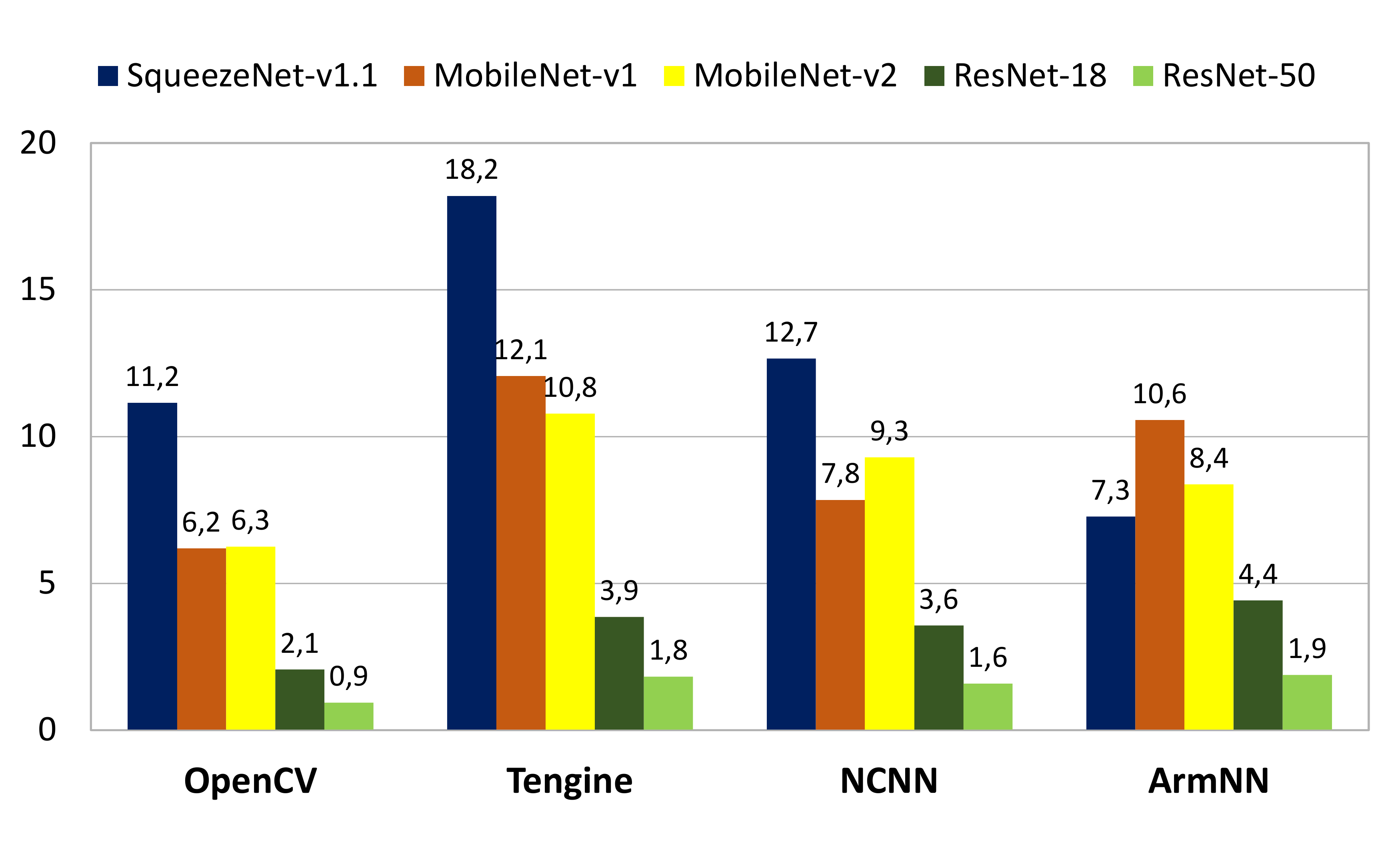}
         \caption{32-bit OS, case (a) -- no cooling}
     \end{subfigure}
          \hfill
     \begin{subfigure}[b]{0.49\textwidth}
         \centering
         \includegraphics[width=\textwidth]{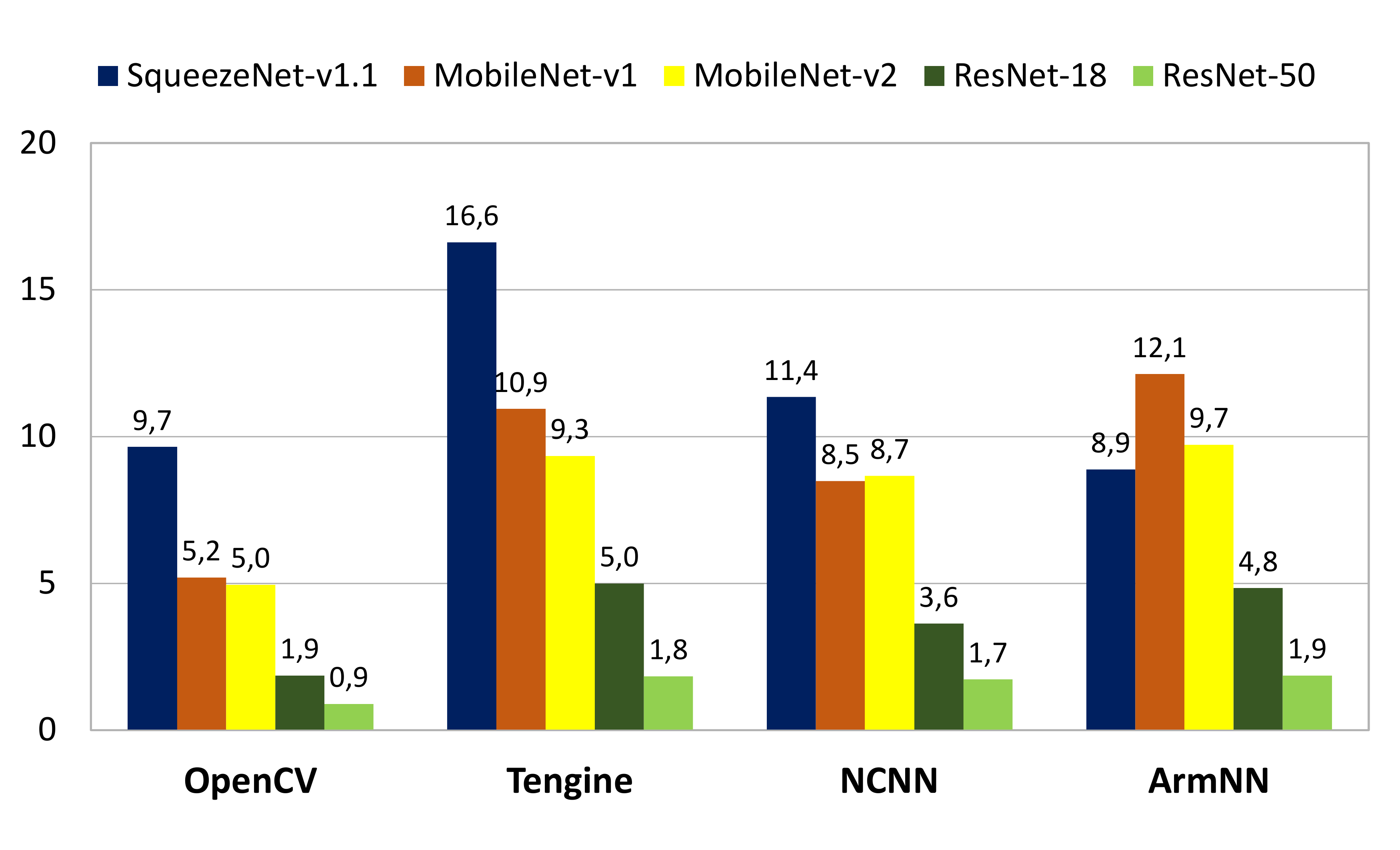}
         \caption{64-bit OS, case (a) -- no cooling}
     \end{subfigure}
     \vfill
     \vspace*{2mm}
     \begin{subfigure}[b]{0.49\textwidth}
         \centering
         \includegraphics[width=\textwidth]{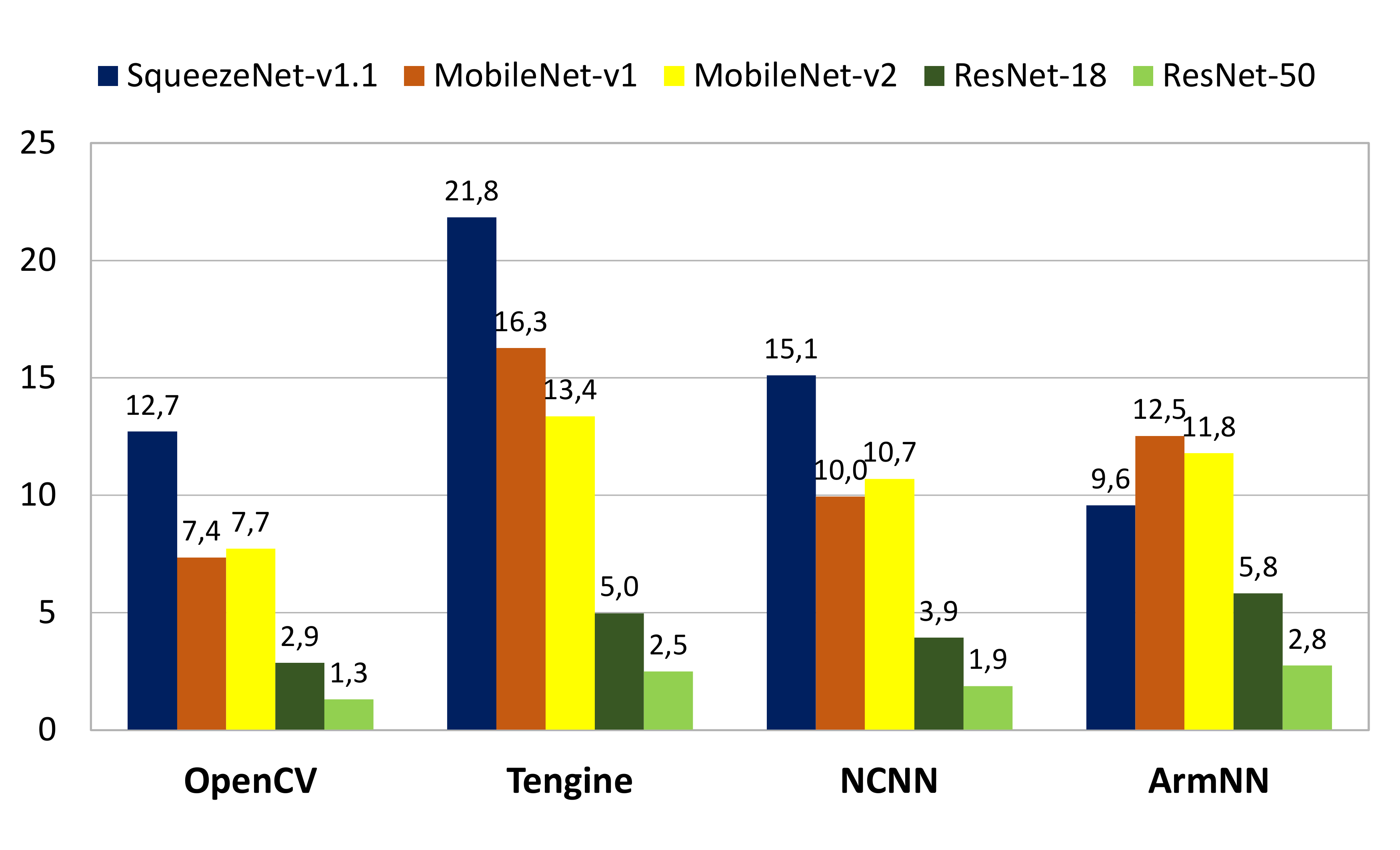}
         \caption{32-bit OS, case (b) -- active cooling}
     \end{subfigure}
     \hfill
     \begin{subfigure}[b]{0.49\textwidth}
         \centering
         \includegraphics[width=\textwidth]{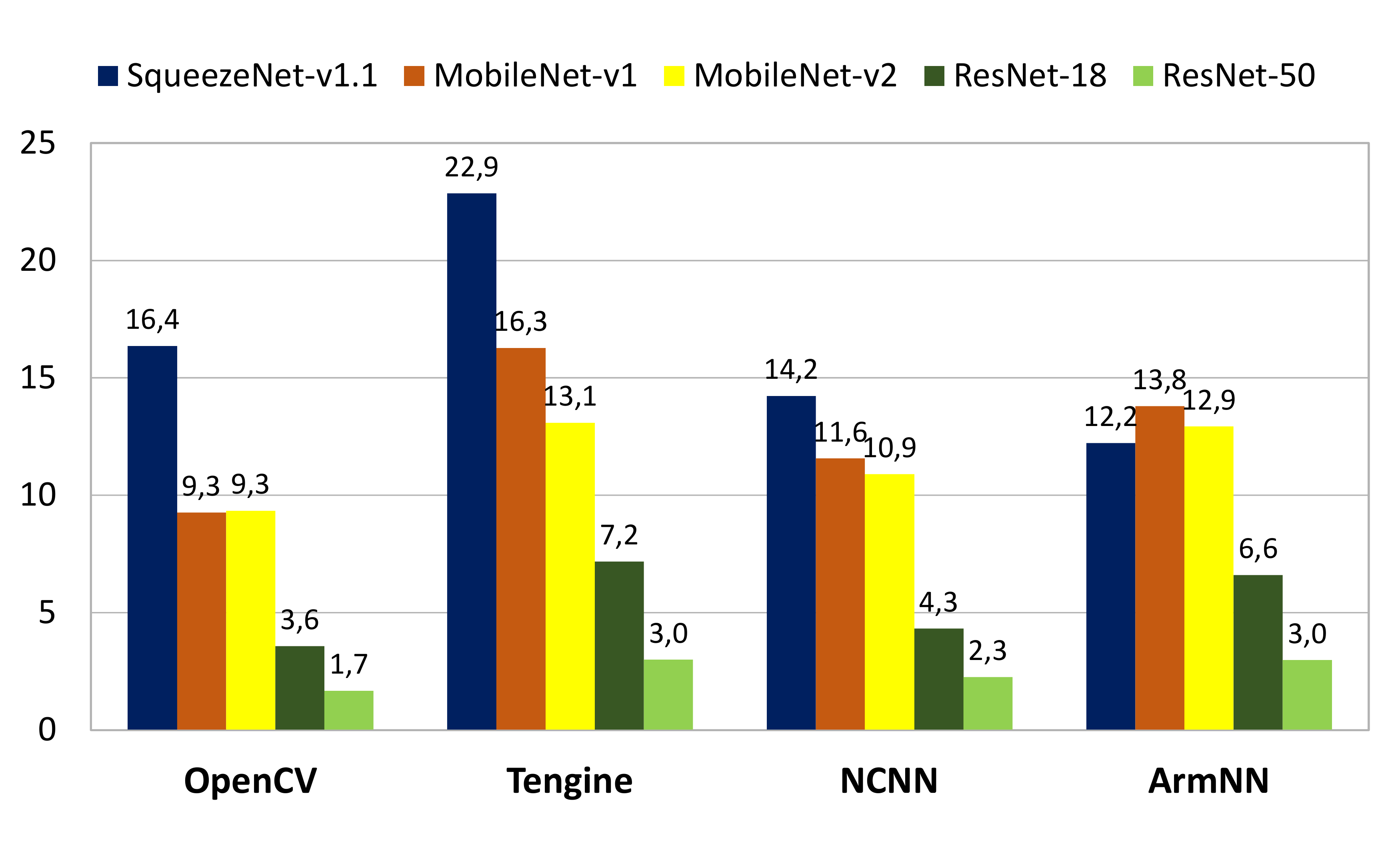}
         \caption{64-bit OS, case (b) -- active cooling}
     \end{subfigure}     
     \vspace*{1mm}
        \caption{Average steady-state throughput (frames per second) for cases (a) -- no cooling -- and (b) -- active cooling.}
        \label{fig:FPS}
\end{figure*}

 Indoor experiments were conducted under stable ambient temperature. Considering every possible combination of the aforementioned OSs, DL frameworks, and CNNs\footnote{ArmNN v20.05 did not support SqueezeNet at the moment of performing these tests. Thus, we specifically employed v20.08 for this CNN.} along with the two general cases, (a) and (b), a total of 80 inference scenarios were characterized. For all of them, the tests were properly extended until reaching steady-state performance. As an example, the longest indoor test lasted 5340 s for 32-bit OS/OpenCV/ResNet-50 with no cooling. The results are summarized in Figs.~\ref{fig:FPS} to~\ref{fig:fan_ratio} and discussed below. Note that Fig.~\ref{fig:CPU_freq} only shows the average steady-state CPU frequency for case (a) given that this parameter always keeps at its maximum (1.5 GHz) when active cooling was applied. Likewise, Fig.~\ref{fig:fan_ratio} only depicts the fan usage for case (b). As a general comment, the variability of these results demonstrate the importance of carefully selecting a set of hardware and software components suitable for the target application. 


\begin{itemize}
\item \textbf{Throughput}. Figure~\ref{fig:FPS} shows the average steady-state throughput (expressed in frames per second) for cases (a) -- no cooling -- and (b) -- active cooling. As expected, there is a clear inverse correlation between the complexity of the models, reflected by the rightmost column in Table~\ref{tab:CNNs}, and the average frame rate: the higher the network complexity, the lower the throughput. As pointed out in Section~\ref{CNNs}, the accuracy does not follow such correlation so markedly. Thus, throughput is not always traded off for a more precise inference. In such cases, the global performance is significantly degraded. Tables \ref{tab:best_fps_without} and \ref{tab:best_fps_cooling} show the best combination of OS and DL framework for each CNN. Remarkably, Tengine achieves the highest throughput in all cases but one (ResNet-50, no cooling), for which anyway Tengine performs closely to ArmNN.  Note that, as expected, active cooling always leads to better performance than no cooling, no matter the combination considered. However, this is not the case for the 64-bit OS versus 32-bit OS. Indeed, with no cooling, many combinations performed better on the 32-bit OS. We conjecture that this is due to the specificity of each framework when it comes to exploiting the underlying system architecture. Finally, note that there are combinations for which active cooling is particularly beneficial. For instance, this is the case of OpenCV on the 64-bit OS, for which active cooling improves the throughput by approximately 82\% on average for the five CNNs, with an absolute maximum of approximately 90\% for ResNet-18. Interestingly, the average improvement for OpenCV on the 32-bit OS is only 27.5\%. 

\begin{table}[b!]
\caption{Combinations achieving maximum throughput for case (a) -- no cooling.}
\label{tab:best_fps_without} 
\centering
\scalebox{0.99}{
\begin{tabular}{l l l}
  \toprule
  \textbf{Network} & \textbf{Software} & \textbf{Throughput}\\
  \midrule
  \textbf{SqueezeNet-v1.1} & 32-bit Tengine &  18.2 fps\\
  \textbf{MobileNet-v1} & 32-bit Tengine / 64-bit ArmNN & 12.1 fps \\
  \textbf{MobileNet-v2} &  32-bit Tengine & 10.8 fps \\
  \textbf{ResNet-18} & 64-bit Tengine & 5.0 fps \\
  \textbf{ResNet-50} & 32/64-bit ArmNN & 1.9 fps \\
  \bottomrule
\end{tabular}}
\end{table}

\begin{table}[b!]
\caption{Combinations achieving maximum throughput for case (b) -- active cooling.}\label{tab:best_fps_cooling} 
\centering
\scalebox{0.99}{
\begin{tabular}{l l l}
  \toprule
  \textbf{Network} & \textbf{Software} & \textbf{Throughput}\\
  \midrule
  \textbf{SqueezeNet-v1.1} & 64-bit Tengine &  22.9 fps\\
  \textbf{MobileNet-v1} & 32/64-bit Tengine &  16.3 fps \\
  \textbf{MobileNet-v2} &  32-bit Tengine &  13.4 fps \\
  \textbf{ResNet-18} & 64-bit Tengine & 7.2 fps \\
  \textbf{ResNet-50} & 64-bit Tengine/ArmNN &  3.0 fps \\
  \bottomrule
\end{tabular}}
\end{table}

\item \textbf{CPU usage}. Figure~\ref{fig:CPU_usage} shows the average and standard deviation of the steady-state CPU usage. This metric is particularly interesting because it provides information about the available CPU power to carry out other tasks. The first aspect to point out is that, generally speaking, the application of active cooling hardly affects the average CPU usage. However, it does have an impact on the standard deviation, in particular for the 64-bit OS, achieving greater stability -- i.e., smaller standard deviation -- at steady state. Note that Tengine, in addition to achieving the highest frame rates, is the most efficient framework in terms of CPU usage for the majority of combinations. By contrast, ArmNN is the less efficient in most of the cases. 

\begin{figure*}[!b]
     \centering
     \begin{subfigure}[b]{0.49\textwidth}
         \centering
         \includegraphics[width=\textwidth]{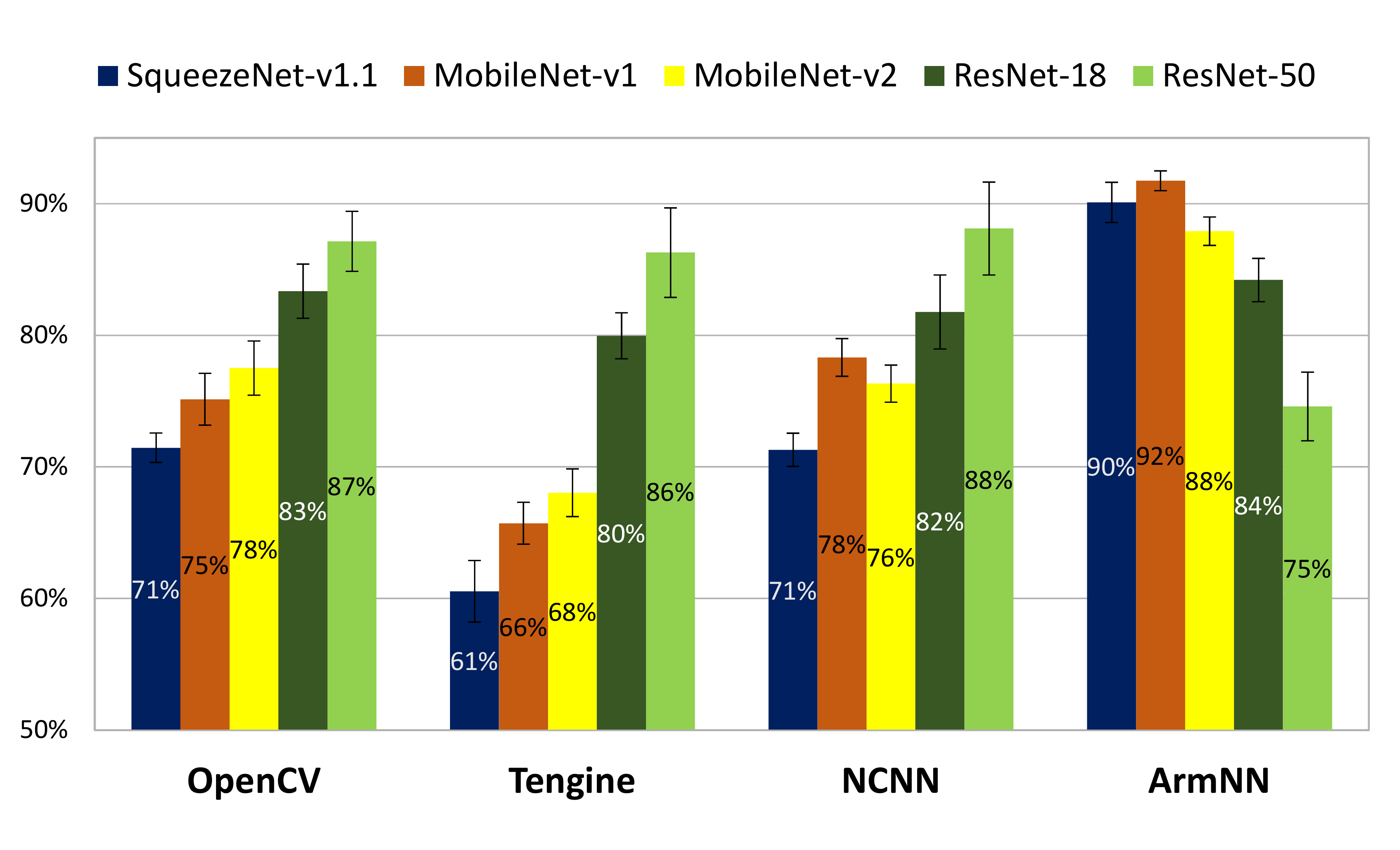}
         \caption{32-bit OS, case (a) -- no cooling}
     \end{subfigure}
    \hfill
     \begin{subfigure}[b]{0.49\textwidth}
         \centering
         \includegraphics[width=\textwidth]{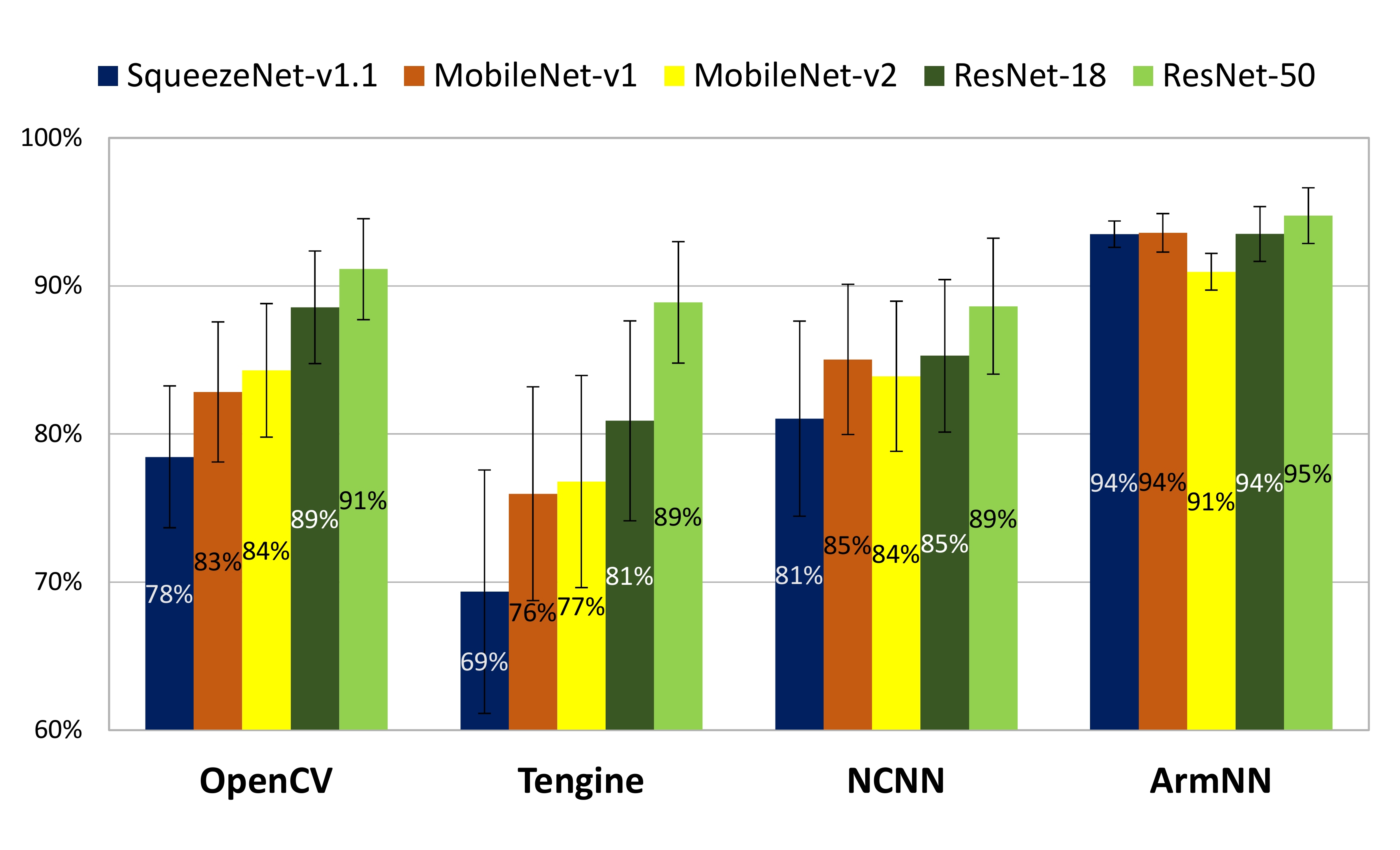}
         \caption{64-bit OS, case (a) -- no cooling}
     \end{subfigure}
     \vfill
     \vspace*{2mm}
     \begin{subfigure}[b]{0.49\textwidth}
         \centering
         \includegraphics[width=\textwidth]{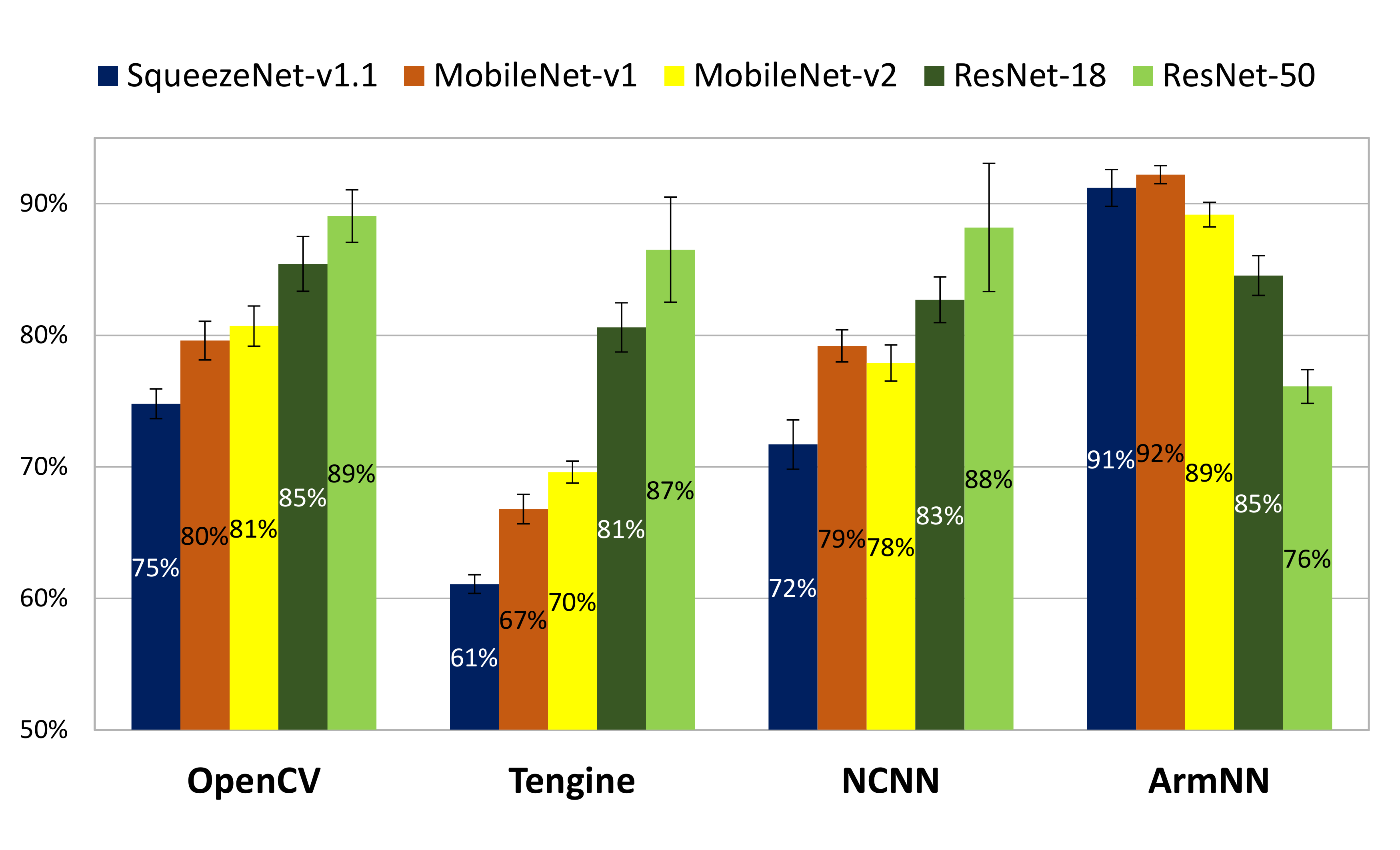}
         \caption{32-bit OS, case (b) -- active cooling}
     \end{subfigure}
     \hfill
    \begin{subfigure}[b]{0.49\textwidth}
         \centering
         \includegraphics[width=\textwidth]{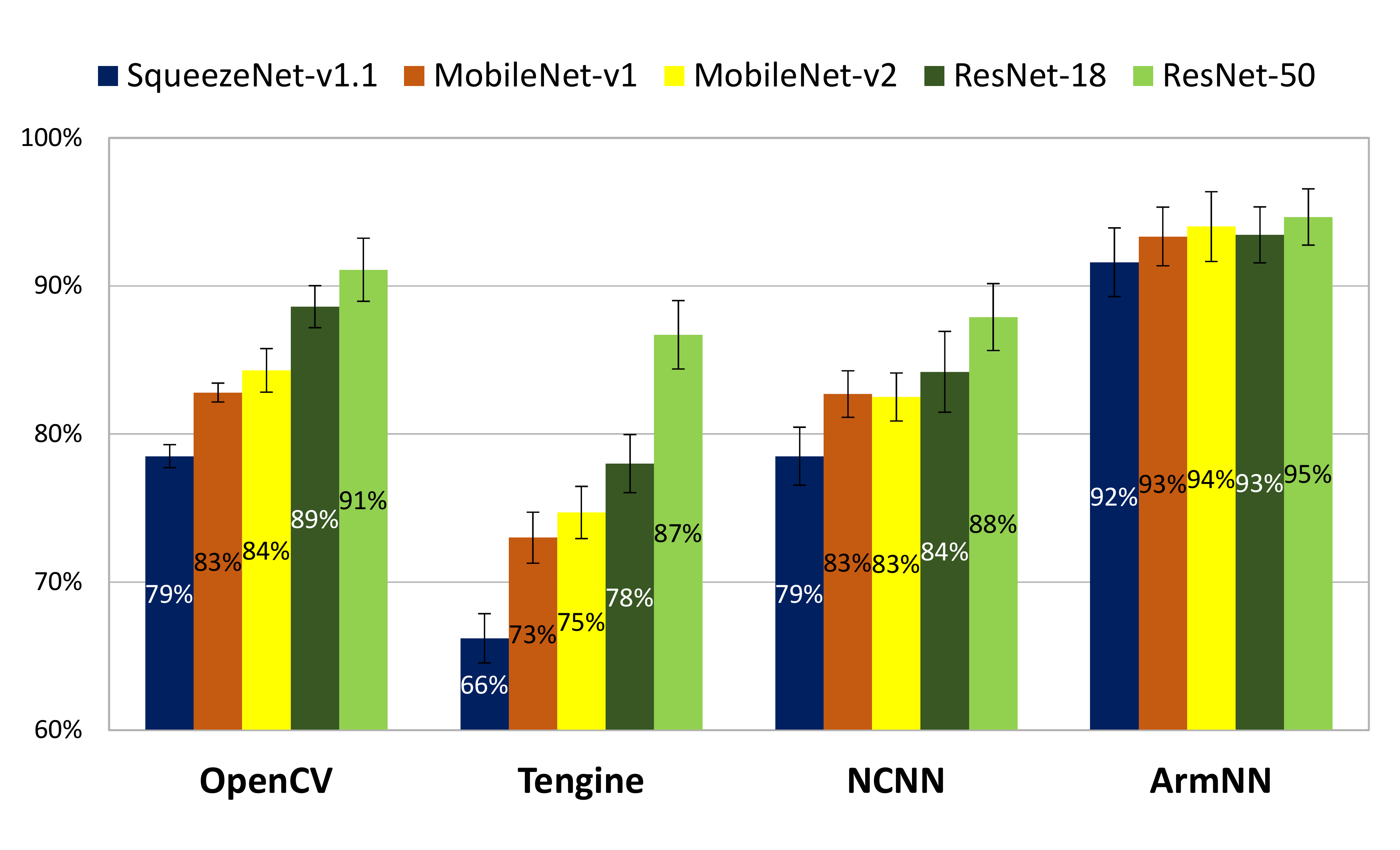}
         \caption{64-bit OS, case (b) -- active cooling}
     \end{subfigure}     
     \vspace*{1mm}
        \caption{Average and standard deviation of steady-state CPU usage for cases (a) -- no cooling -- and (b) -- active cooling.}
       \label{fig:CPU_usage}
\end{figure*}

\item \textbf{CPU frequency}. The average CPU frequency\footnote{The fact that we calculate the average explains the granularity of values of the CPU frequency, which actually takes instantaneous values from a reduced set within the interval [600 MHz, 1.5 GHz].} for case (a) -- no cooling -- is shown in Fig.~\ref{fig:CPU_freq}. This metric is directly related with power consumption. For the 32-bit OS, ArmNN operates at the lowest frequency for four out of the five network models whereas OpenCV runs at the highest frequency also for four of the CNNs. Interestingly, this changes for the 64-bit OS, on which OpenCV operates at the lowest CPU frequency for four models and NCNN clearly reaches the highest frequency in all cases. Again, keep in mind that for case (b) -- active cooling --, the CPU frequency never decreased from its maximum value, i.e., 1.5 GHz.  

\item \textbf{Fan usage}. Figure~\ref{fig:fan_ratio} shows the values of $FU (\%)$ for case (b) -- active cooling. This parameter determines the extra consumption required for precluding thermal throttling. Note that NCNN is the framework demanding the lowest usage of active cooling in most of the cases for both the 32-bit and 64-bit OSs. By contrast, ArmNN is clearly the framework requiring the longest time of fan for the 32-bit OS. Concerning the 64-bit OS, OpenCV presents high fan usage -- not necessarily the highest -- for the five CNN models. Globally, the values of $FU (\%)$ range from 33\% to 65\%.

\begin{figure*}[!b]
     \centering
     \begin{subfigure}[b]{0.49\textwidth}
         \centering
         \includegraphics[width=\textwidth]{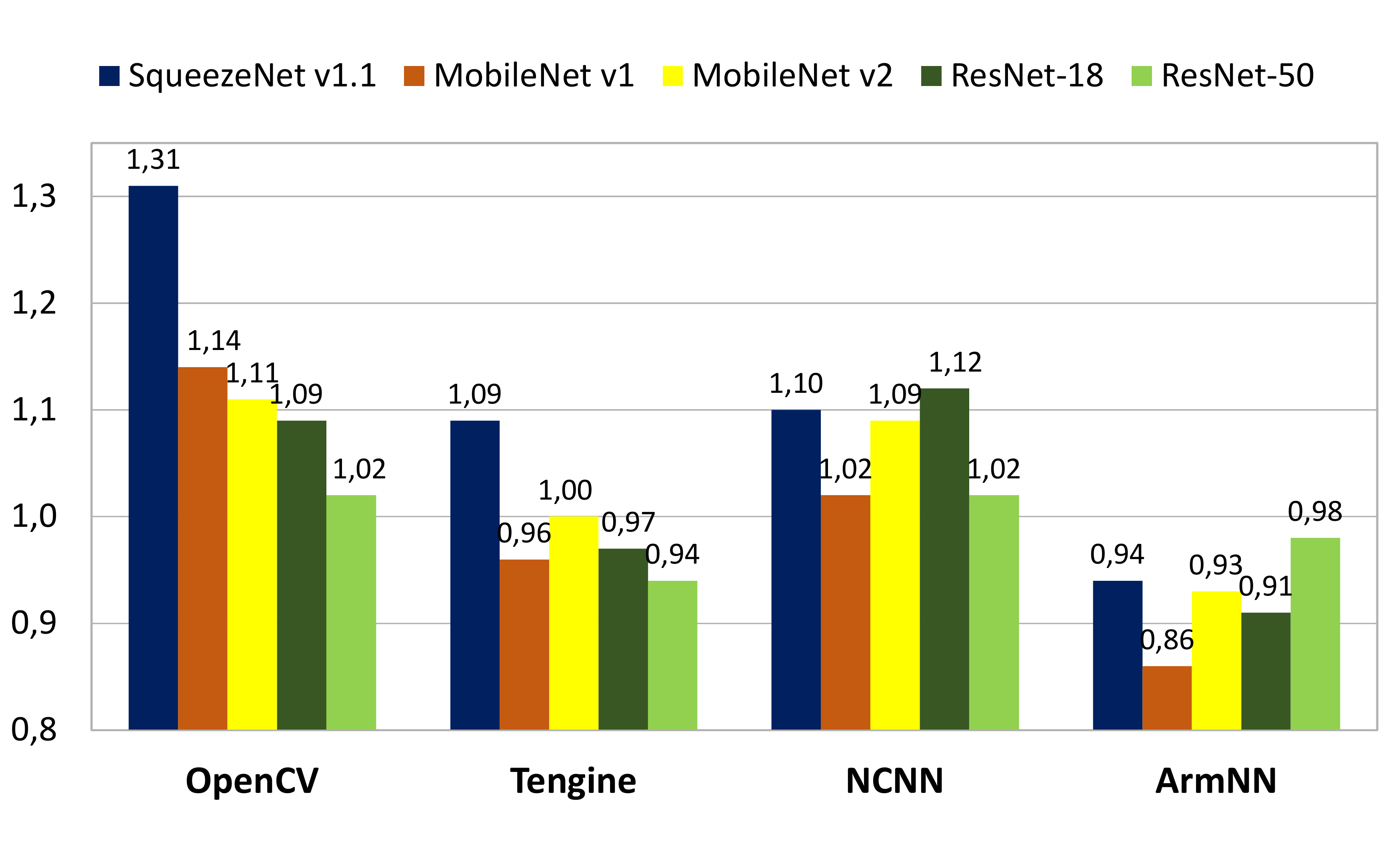}
         \caption{32-bit OS}
     \end{subfigure}
     \vspace*{2mm}
     \begin{subfigure}[b]{0.49\textwidth}
         \centering
         \includegraphics[width=\textwidth]{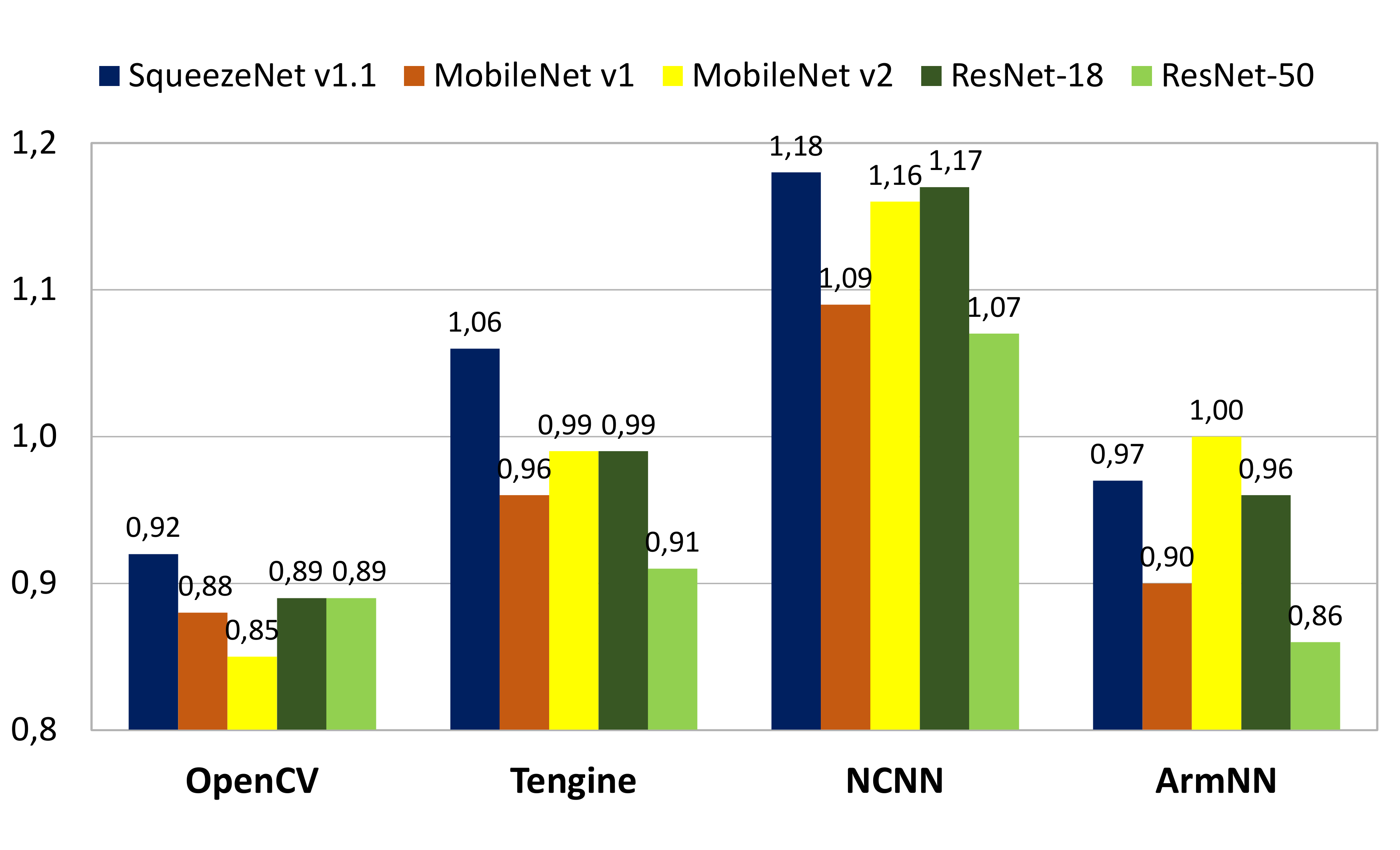}
         \caption{64-bit OS}
     \end{subfigure}
    \vspace*{1mm}
        \caption{Average steady-state CPU frequency (GHz) for case (a) -- no cooling.}
        \label{fig:CPU_freq}
\end{figure*}

\begin{figure*}[!b]
     \centering
     \begin{subfigure}[b]{0.49\textwidth}
         \centering
         \includegraphics[width=\textwidth]{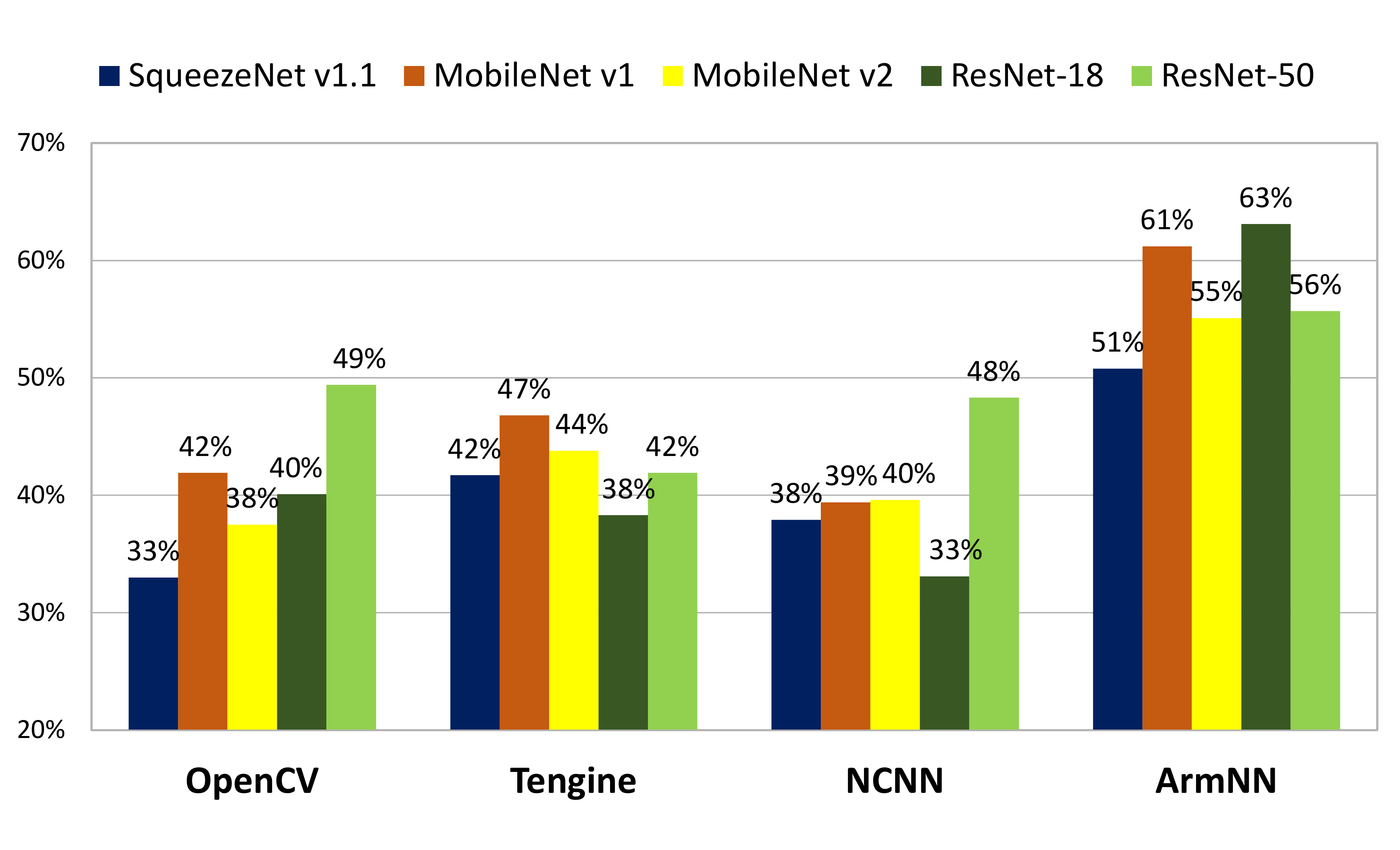}
         \caption{32-bit OS}
     \end{subfigure}
     \vspace*{2mm}
     \begin{subfigure}[b]{0.49\textwidth}
         \centering
         \includegraphics[width=\textwidth]{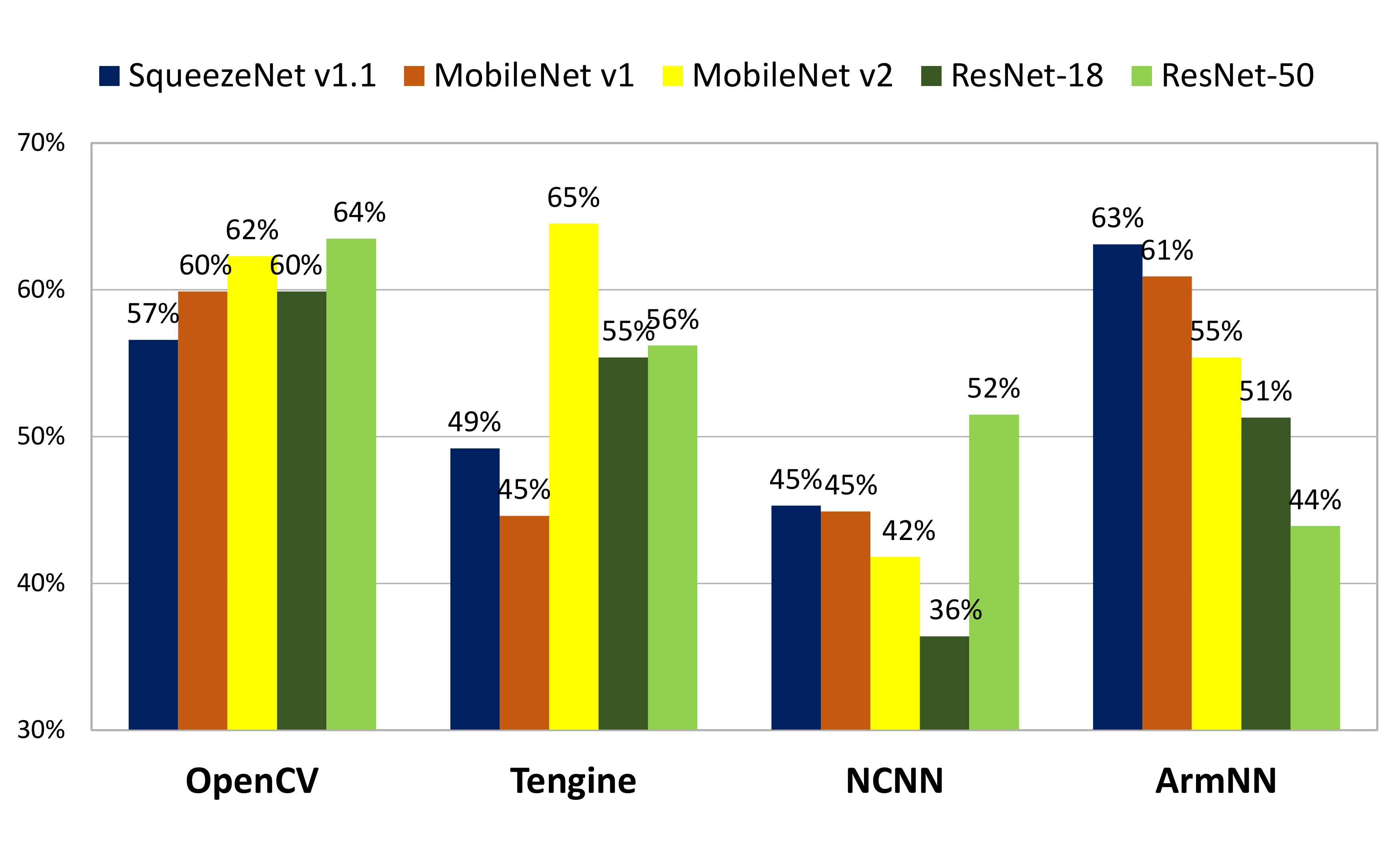}
         \caption{64-bit OS}
     \end{subfigure}
      \vspace*{1mm}
        \caption{Fan usage, $FU (\%)$ according to Eq.~\ref{eq:fan}, for case (b) -- active cooling.}
      \label{fig:fan_ratio}
\end{figure*}

Two important conclusions can be drawn from the above discussions. First, active cooling is worth it as long as its power consumption is affordable. Note that the extra consumption with respect to no cooling stems from two sources: i) the consumption associated with the fan to prevent thermal throttling; ii) the higher consumption derived from the fact the CPU constantly operates at its highest frequency. If affordable, active cooling always improves the throughput, with a significant increase in some cases. Second, Tengine seems to be the best option, no matter the OS version or whether active cooling is applied. This framework achieves the best throughput at the lowest CPU utilization in most scenarios, with moderate CPU frequency in case of no cooling, and moderate fan usage in case of active cooling. 

\end{itemize}

\section{Outdoor Tests: Results and Analysis}
\label{sec:temp_experiment}

 Outdoor visual inference typically suffers from unregulated ambient temperatures that can affect the processor performance. According to the results presented in the previous section, thermal throttling can be prevented through active cooling. However, there are power budgets in outdoor application scenarios -- e.g., remote environmental monitoring -- that could not afford the extra consumption of the fan. Thus, we conducted a number of outdoor tests to evaluate the impact of ambient temperature sweeping a wide variation interval while performing continuous CNN processing on the RPi4B with no active cooling. 

\begin{figure}[!h]
\centering
\includegraphics[width=0.4\textwidth]{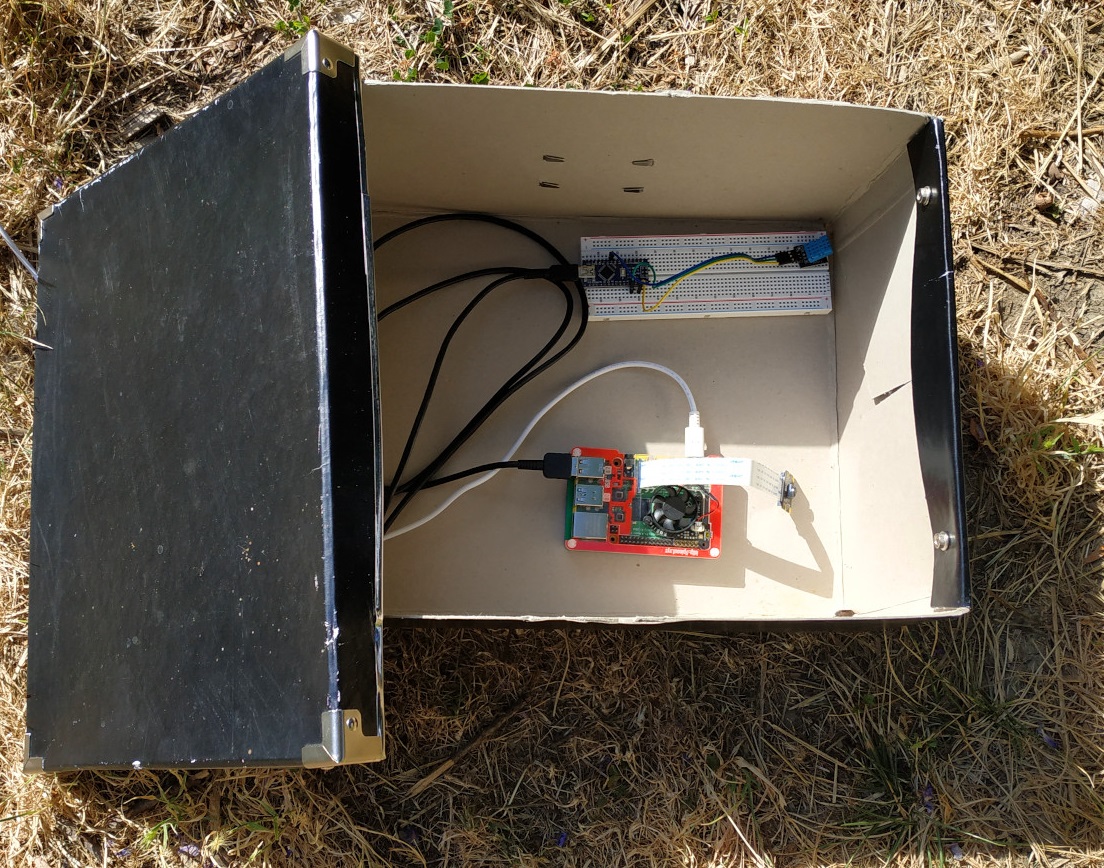}
\caption{Experimental setup for outdoor tests. The whole system was placed in a black cardboard box for protection and temperature homogenization.} 
\label{fig:setpu_temperature}
\end{figure}

The setup for outdoor tests is shown in Fig. \ref{fig:setpu_temperature}. The whole system was placed in a black cardboard box for protection and homogenization of the temperature of all the components. Following the same procedure described in Section~\ref{sec:methodology}, we periodically monitored the instantaneous CNN runtime, the CPU temperature and frequency, and the ambient temperature.  

\begin{figure}[!t]
\centering
\includegraphics[width=0.6\textwidth]{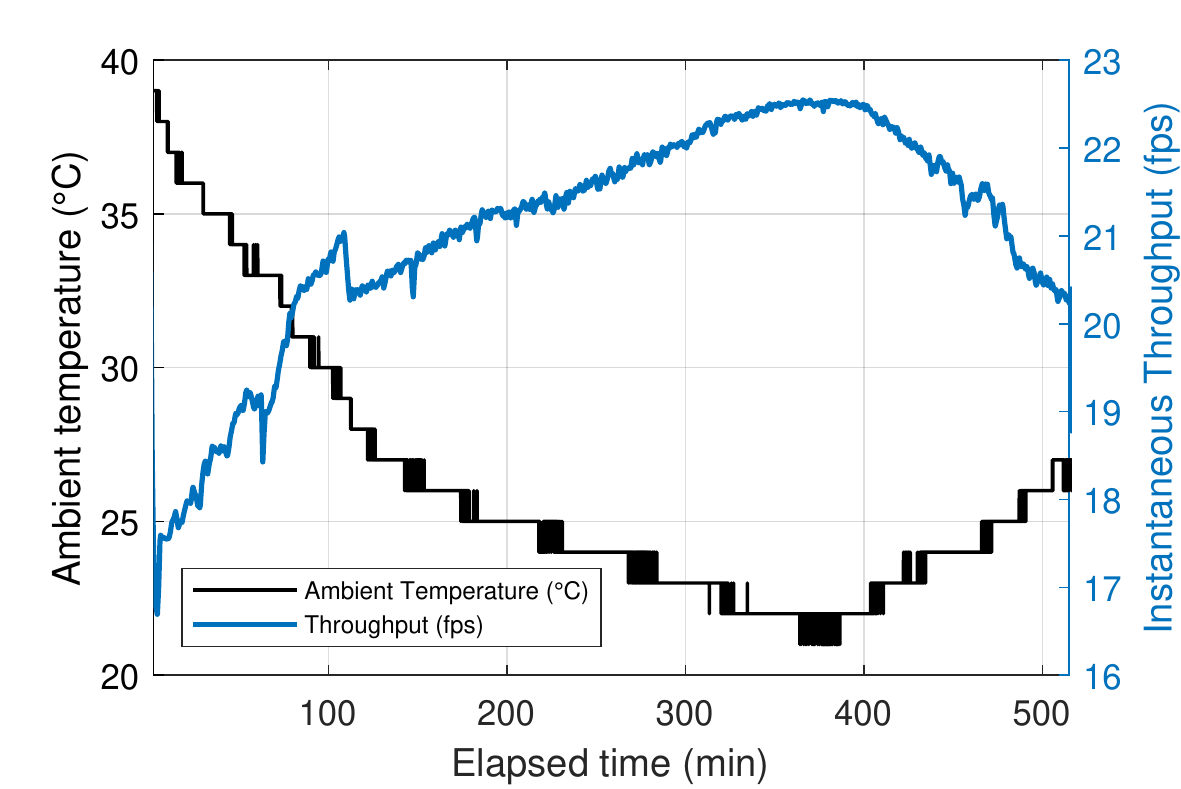}
\caption{Example of temporal evolution of throughput and simultaneously monitored ambient temperature for the 32-bit-OS/Tengine/SqueezeNet combination in outdoor test with no cooling.} \label{fig:ambient_temp}
\end{figure}

\begin{figure}[!h]
\centering
\includegraphics[width=0.6\textwidth]{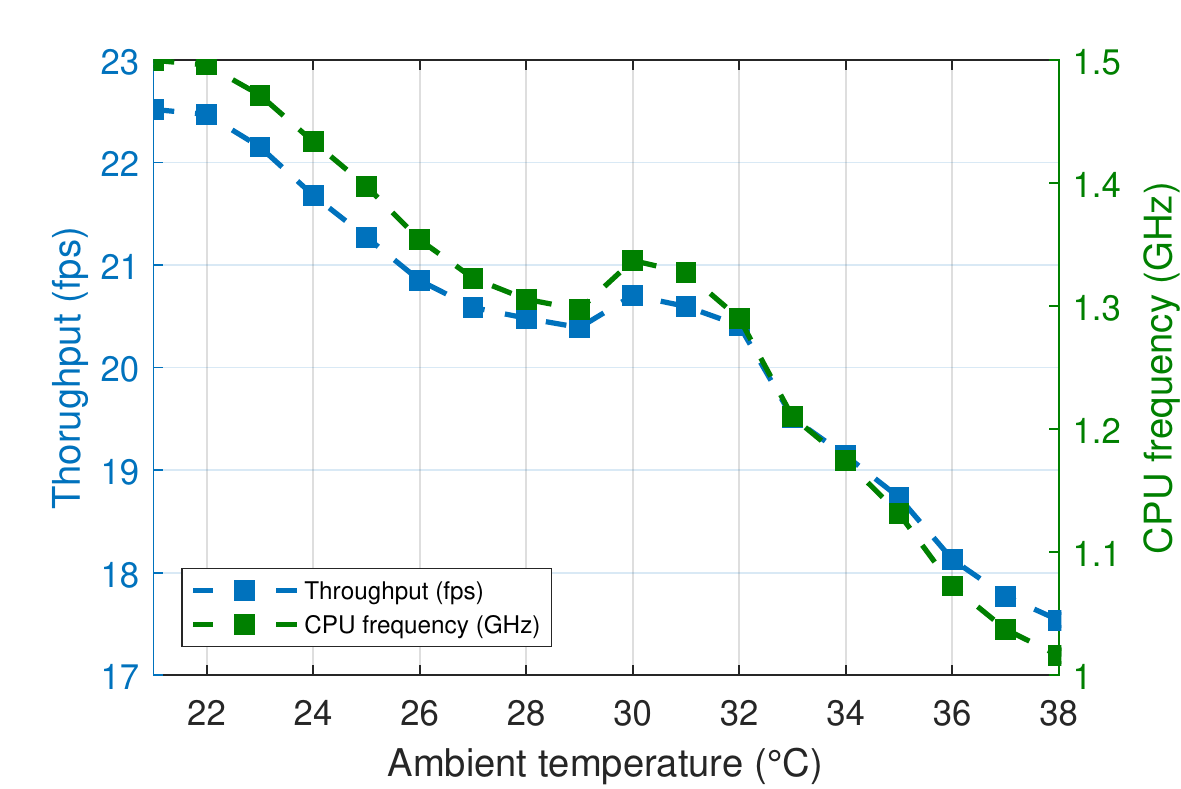}
\caption{Throughput and CPU frequency vs. ambient temperature for the outdoor test shown in Fig.~\ref{fig:ambient_temp}.}
\label{fig:measurements_vs_temp}
\end{figure}

Fig.~\ref{fig:ambient_temp} shows an example of collected temporal data. It corresponds to approximately 520 minutes of continuous inference with 32-bit OS/Tengine/SqueezeNet. In this particular test, the ambient temperature inside the box ranged from 38 \degree C in direct sun in the afternoon to 21 \degree C at night. The throughput correspondingly varied from 17.7 to 22.5 fps. To better understand this temperature dependency, Fig. \ref{fig:measurements_vs_temp} depicts throughput and CPU frequency as a function of the ambient temperature. Data in y-axis represent the average of all the measurements taken for each step of 1\degree C (x-axis). The correlation is evident: higher ambient temperature gives rise to worse inference performance (blue dots) due to thermal throttling expressed in terms of CPU frequency (green dots). 

The proposed metric to analyze the impact of ambient temperature is the \textit{maximum relative throughput variation}, denoted by $\Delta Th (\%)$. It measures the loss of throughput with respect to the maximum value recorded during a particular experiment, i.e., ${Th}_{max}$. Thus, $\Delta Th (\%)$ is expressed as follows:
 \begin{equation}
   \Delta Th (\%) = \frac{  Th_{max} - Th_{min}  }{Th_{max}} \times 100
  \label{eq:fps_variation}
 \end{equation}
where $[Th_{min},Th_{max}]$ represents the range of throughput variation during the corresponding outdoor test.



%

\begin{figure}[!t]
\centering
\includegraphics[width=0.65\textwidth]{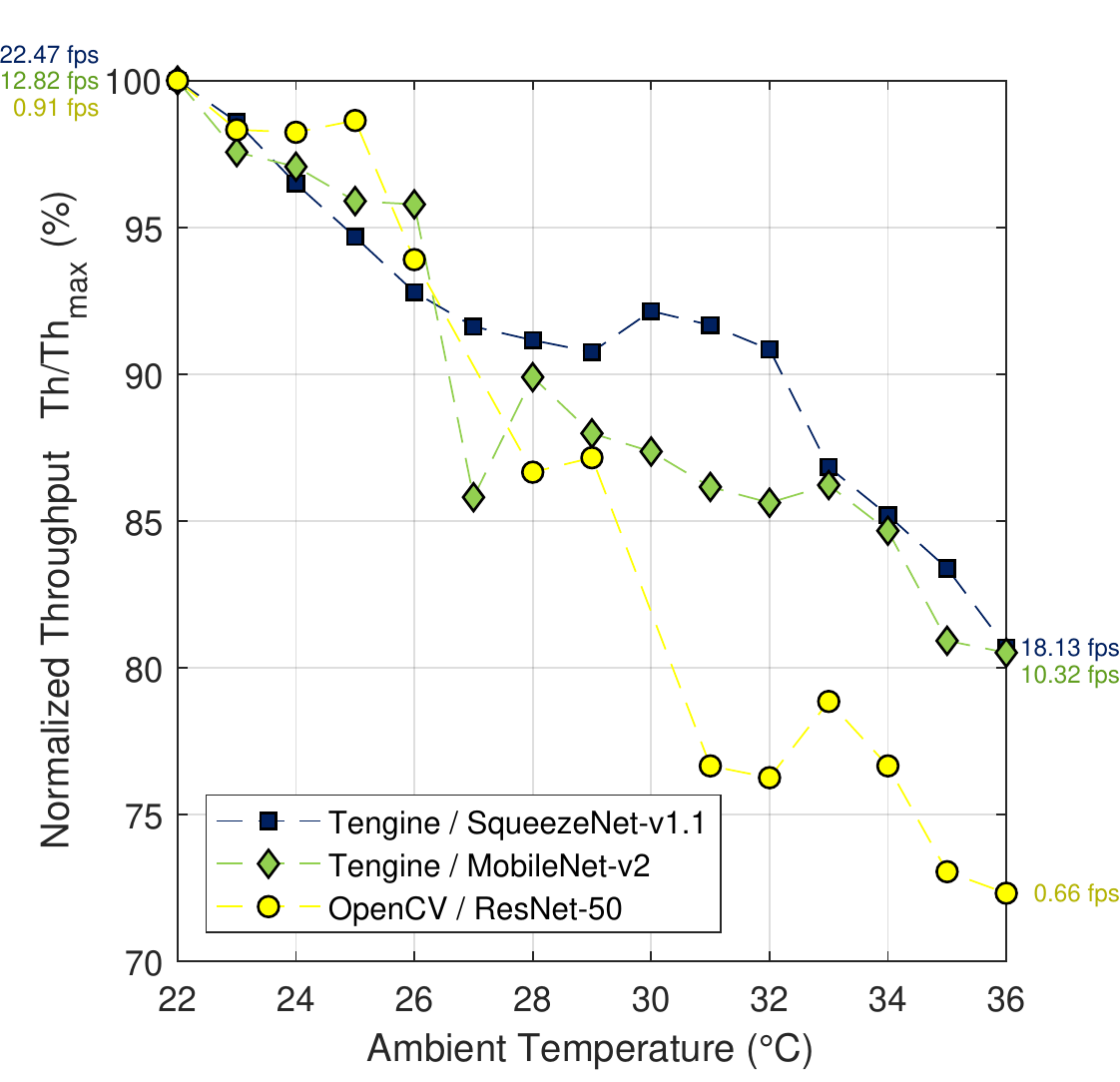}
\caption{Average throughput normalized with respect to $Th_{max}$ as a function of the ambient temperature. The maximum and minimum values of throughput, which occurred at 22 \degree C and 36 \degree C respectively, are also displayed for each curve.} \label{fig:fps_vs_temp}
\end{figure}

Note that the conducted outdoor tests were time-consuming -- typically a few hours were required for ambient temperature to vary in a wide interval -- and strongly dependent of external thermal conditions -- it was difficult to replicate experiments sweeping the same range of temperatures. Eventually, we were able to successfully complete tests for three representative combinations of framework/model on the 32-bit OS: Tengine/SqueezeNet, Tengine/MobileNet-v2, and OpenCV/ResNet-50. This covers from a case featuring low CPU usage and high throughput, i.e., Tengine/SqueezeNet, to a combination characterized by a high computational load such as OpenCV/ResNet-50, with Tengine/MobileNet-v2 as an intermediate case. For these combinations, we were able to reproduce similar conditions, with the ambient temperature always in the range 22 \degree C -- 36 \degree C. Fig. \ref{fig:fps_vs_temp} summarizes the results. It shows the average throughput for each temperature step normalized with respect to its maximum, that is, $Th_{max}$. This maximum occurred at 22 \degree C in all cases; the particular values of throughput at this point are included in the plot. Likewise, the minimum occurred at 36 \degree C in all cases; the particular values of throughput at that point are also included in the plot. The corresponding values of $\Delta Th (\%)$ are reported in Table \ref{tab:Temp_Impactv2}. Remarkably, the impact of varying ambient conditions on continuous inference is not negligible at all. Even for the lightest model (SqueezeNet), $\Delta Th (\%)$ takes a value of 19.3 \%. ResNet-50 on OpenCV is even more sensitive to the ambient temperature, with $\Delta Th (\%)$ reaching 27.7 \%. Depending on the requirements at application level, this variation could be critical to continuously meet the expected performance in real operation conditions. 

\begin{table}[h!]
\caption{Values of $\Delta Th (\%)$ (defined in Eq. \ref{eq:fps_variation}) for the three combinations of outdoor tests conducted.}\label{tab:Temp_Impactv2}
\centering
\scalebox{0.99}{
\begin{tabular}{l c}
  \toprule
  \textbf{Framework / Network } & \textbf{ {$\Delta Th$} }\\
  \midrule
  Tengine / SqueezeNet-v1.1 & $19.3 \%$\\
  Tengine / MobileNet-v2  & $19.5 \%$\\
  OpenCV / ResNet-50 & $27.7 \%$\\
  \bottomrule
\end{tabular}}
\end{table}
 

\section{Conclusions}
\label{sec:conclusion}

In this study, we demonstrate that thermal effects cannot be neglected when it comes to designing embedded vision systems for real application scenarios. Thermal throttling can notably decrease the throughput during long-term continuous inference, with further degradation in case of high ambient temperature. Active cooling can prevent it, but at the cost of extra power consumption. In both cases -- no cooling and active cooling --, the resulting performance varies significantly depending on the OS, software framework, and CNN model. Thus, benchmarking and tests conducted under real operation conditions become fundamental to gain reliable insight about the expected behavior. 
Future work will address the development of techniques to maximize the performance of embedded systems tailored for remote sensing in terms of throughput, accuracy, and battery lifetime. These techniques will cover from advanced temperature and power managing algorithms to dynamic adaptation of the image resolution according to the scene content to optimize the use of the available computational power.

\section*{Acknowledgements}
This work was supported by Spanish Government MICINN (European Region Development Fund, ERDF/FEDER) through project RTI2018-097088-B-C31, European Union H2020 MSCA through project ACHIEVE-ITN (Grant No. 765866), and by the US Office of Naval Research through Grant No. N00014-19-1-2156. 

\bibliographystyle{IEEEtran} 
\bibliography{bibliography}

\end{document}